\documentclass{article}

\usepackage{microtype}
\usepackage{graphicx}
\usepackage{subfigure}
\usepackage{booktabs} 
\usepackage{hyperref}

\usepackage[accepted]{icml2024}

\usepackage{amsmath}
\usepackage{amssymb}
\usepackage{mathtools}
\usepackage{amsthm}

\usepackage[capitalize,noabbrev]{cleveref}

\theoremstyle{plain}

\theoremstyle{definition}

\theoremstyle{remark}

\usepackage{amsthm}
\usepackage{amsmath}
\usepackage{enumitem}
\usepackage{booktabs}
\usepackage{xcolor}
\definecolor{mygrey}{cmyk}{0, 0, 0, 0.5}
\newcommand{\smallgrey}[1]{\textcolor{mygrey}{\scalebox{0.8}{#1}}}
\usepackage{multirow}
\usepackage{amssymb}
\usepackage{bm}
\usepackage{csquotes}

\usepackage[textsize=tiny]{todonotes}

\icmltitlerunning{Jacobian Regularizer-based Neural Granger Causality}

\begin{document}

\twocolumn[

\icmltitle{Jacobian Regularizer-based Neural Granger Causality}

\icmlsetsymbol{equal}{*}

\begin{icmlauthorlist}
\icmlauthor{Wanqi Zhou}{xjtu,riken}
\icmlauthor{Shuanghao Bai}{xjtu}
\icmlauthor{Shujian Yu}{vua}
\icmlauthor{Qibin Zhao}{riken}
\icmlauthor{Badong Chen}{xjtu}
\end{icmlauthorlist}

\icmlaffiliation{xjtu}{Institute of Artificial Intelligence and Robotics, Xi’an Jiaotong University, China.}
\icmlaffiliation{riken}{RIKEN AIP, Japan.}
\icmlaffiliation{vua}{Vrije Universiteit Amsterdam, Netherlands}
\icmlcorrespondingauthor{Badong Chen}{chenbd@mail.xjtu.edu.cn}

\icmlkeywords{Machine Learning, ICML}

\vskip 0.3in
]


\printAffiliationsAndNotice{} 

\begin{abstract}

With the advancement of neural networks, diverse methods for neural Granger causality have emerged, which demonstrate proficiency in handling complex data, and nonlinear relationships.
However, the existing framework of neural Granger causality has several limitations. It requires the construction of separate predictive models for each target variable, and the relationship depends on the sparsity on the weights of the first layer, resulting in challenges in effectively modeling complex relationships between variables as well as unsatisfied estimation accuracy of Granger causality.
Moreover, most of them cannot grasp full-time Granger causality.
To address these drawbacks, we propose a \textbf{J}acobian \textbf{R}egularizer-based \textbf{N}eural \textbf{G}ranger \textbf{C}ausality (\textbf{JRNGC}) approach, a straightforward yet highly effective method for learning multivariate summary Granger causality and full-time Granger causality by constructing a single model for all target variables. 
Specifically, our method eliminates the sparsity constraints of weights by leveraging an input-output Jacobian matrix regularizer, which can be subsequently represented as the weighted causal matrix in the post-hoc analysis.
Extensive experiments show that our proposed approach achieves competitive performance with the state-of-the-art methods for learning summary Granger causality and full-time Granger causality while maintaining lower model complexity and high scalability. 

\end{abstract}

\section{Introduction}
\label{intro}

\begin{figure*}[ht]
	\centering
	\includegraphics[width=0.75\textwidth]{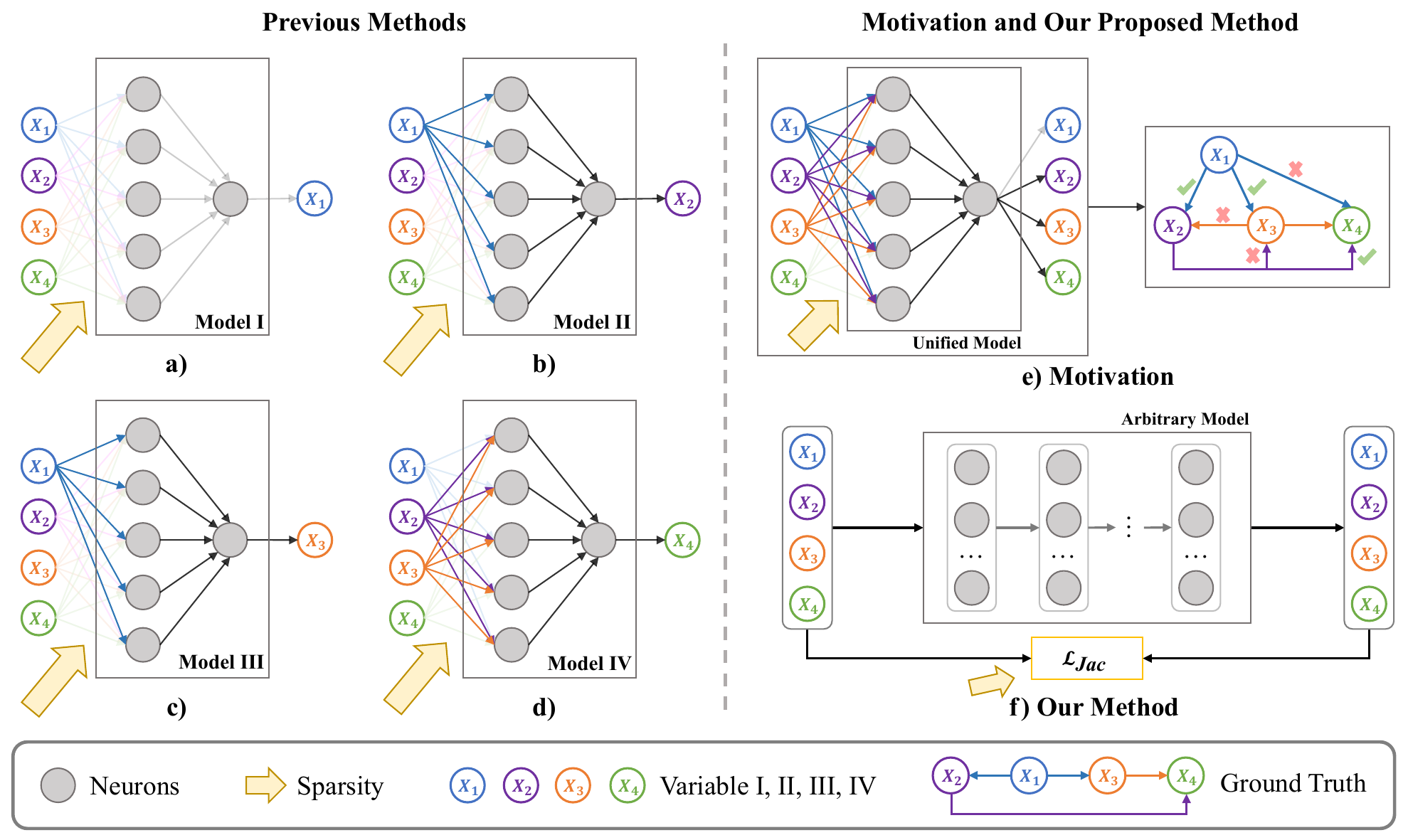}
	\caption{
                Motivation and our proposed neural Granger causality method.
                To learn the true Granger causality, we need to estimate the importance of a variable in helping forecast another variable.
		      For example, let's examine a simple scenario involving the summary causal relationship: $X_4\leftarrow X_2 \leftarrow X_1\rightarrow X_3\rightarrow X_4$. 
		To comprehend this relationship, current neural Granger causality methods need to construct and train the same number of models as the dimensions of the variables to disentangle the importance of each variable and obtain Granger causality by incorporating sparsity penalties on the first layer of each model, as illustrated in figures a)-d).
	However, sparse first-layer network parameters will result in challenges in effectively modeling complex relationships between variables as well as unsatisfied estimation accuracy of Granger causality.
        In addition, it will get wrong Granger causality if we use a multivariate time series forecasting model because of the existence of a shared hidden layer, as exemplified in figure e).
		Instead, our method only needs to build and train a single multivariate time series forecasting model by introducing input-output Jacobian regularizer $\mathcal{L}_{J a c}$. Note that the numbers I, II, III, and IV mean the same model architecture with independent training for different variables and we used two-layer Perceptron for the convenience of illustration.}
	\label{fig:model_figure}
\end{figure*}

On time-series data, causal relations can be learned by various methodologies, such as Granger causality \cite{granger1969investigating}, transfer entropy (TE) \cite{schreiber2000measuring}, 
and other approaches like constraint-based methods (e.g., PCMCI), noise-based methods (e.g., VarLiNGAM), and more, as discussed in the literature \cite{assaad2022survey}.
Traditionally, methodologies like conditional TE \cite{sun2015causal,zhou2022causality} and constraint-based algorithms such as PCMCI, tsFCI \cite{runge2019detecting,entner2010causal} have been employed for uncovering nonlinear causal relationships. 
These methods have become popular due to their capacity to sidestep the necessity for parametric modeling.
However, they grapple with a significant challenge when it comes to scaling up to large causal graphs because of their iterative nature and dependence on the estimation of probability densities, which motivates the need for more efficient and effective techniques in causal relationship identification within time series datasets.

Recent advancements in deep learning have shown promising progress in learning Granger causality. Several studies have explored the application of neural networks, such as multi-layer perceptron (MLP), recurrent neural networks (RNNs), and dynamic variational autoencoders, for learning Granger causality from time series data \cite{tank2021neural, suryadi2023granger, bussmann2021neural, Li_Yu_Principe_2023}. These methods integrate Granger causality as a null hypothesis into time series forecasting or generative models, allowing causal relationships between variables to be more accurately captured through causal constraint modeling.
Despite these advancements, current neural Granger causality methods have two major drawbacks.
Firstly, as illustrated in the left panel of Figure~\ref{fig:model_figure}, they require building separate neural networks for each target variable during the training phase to learn Granger causality.
This can result in computational inefficiencies and scalability challenges when dealing with larger and more complex datasets, especially when we need to grasp long full-time Granger causality.
Secondly, they rely on sparse constraints on the first layer weights, which leads to the necessity of separate models for target variables, and also limits the ability to capture Granger causal relationships for complex data, because the dependencies between variables over time will be overlooked.
Moreover, in cases where the number of variables is high, the search space for causal variance becomes broad.
Thus solely constraining the first layer is insufficient and may result in local optima.
In other words, the relationship between predicting variables and learning causal connections is mutually influential; accurate predictions facilitate genuine causal learning, and, conversely, a proficient understanding of causal relationships enhances predictive accuracy.
Furthermore, these methods render RNNs (e.g., RNN, LSTM) incapable of learning full-time Granger causality, because in a standard RNN,  different time steps of the same time series share the same weight parameters, thus preventing the model from capturing different Granger causality across time steps. 
This limitation compromises the full realization of neural networks' potential in certain applications.

For the first issue, to develop a unified neural Granger causality framework with a multivariate time series forecasting model, it is essential to mitigate the impact of shared hidden layers. 
In the context of a multivariate time series forecasting model, shared hidden layers would obtain the information from all variables and send it to each target variable, leading to the difficulty of disentangling the importance of each variable.  
Thus we must identify a suitable Granger causality estimation method capable of expressing the importance of one variable to another.
For the second issue, we need to find a global Granger causality constraint method.
To solve these issues, we propose a novel approach termed the \textbf{J}acobian \textbf{R}egularization-based \textbf{N}eural \textbf{G}ranger \textbf{C}ausality (JRNGC).
For a detailed view, see the right panel in Figure \ref{fig:model_figure}.
Firstly, the input-output Jacobian matrix can represent the relations between variables over time and does not purely rely on the first layer of the neural network. By imposing a sparse constraint on it, we can explicitly set the null hypothesis that there is no Granger causality between certain variables.
Therefore, the input-output Jacobian matrix can handle the above-mentioned two major drawbacks of NGC.
Furthermore, we utilize the residual MLP neural network, proven more capable of predicting time series \cite{das2023longterm,Zeng_Chen_Zhang_Xu_2023}, although our framework is adaptable to any model.

Our main contributions to this work are as follows:

\begin{itemize}
	\item  To our best knowledge, this is the first work to harness a single NN model with shared hidden layers for multivariate Granger causality analysis. 
        \item We propose a novel neural network framework to learn  Granger causality by incorporating an input-output Jacobian regularizer in the training objective. 
	\item Our method can not only obtain the summary Granger causality but also the full-time Granger causality.
	\item We evaluate our method on commonly used benchmark datasets with extensive experiments.
	Our method can outperform state-of-the-art baselines and show an excellent ability to discover Granger causality, especially for sparse causality.
	
\end{itemize}

\section{Background and Related Work}

\subsection{Background: Granger causality}


Granger causality \cite{granger1969investigating} is widely used for learning causality from time-series data. 
It is a statistical concept of causality that is based on prediction.
It follows Wiener's theory \cite{wiener1956theory}: if the prediction of variable $B$ can be improved by incorporating the past information of variable $A$, along with its own past information, then $A$ Granger causes
$B$.
Given a multivariate time series $\mathbf{x}=\{x_1,x_2,\cdots,x_D\}$ with $D$ dimensions, we model the time series as:
\begin{equation}
    x_j(t)=f_j\left(x_1(<t), \ldots, x_D(<t)\right)+\varepsilon_j,
\label{background}
\end{equation}
where the $\varepsilon_j$ is an independent noise item, and $x_i(<t)$ denotes the past information of time series $x_i$.
The prediction of the value of $x_j$ at time $t$ depends on the past information of other time series, which are the potential causes or \enquote{parent} of $x_j$. 
If a time series $x_i$ (at times before $t$, i.e., $x_i(<t)$) significantly improves the accuracy of predicting the future values of $x_j$ when added to the set of \enquote{parent} time series, then $x_i$ is said to have Granger causality on $x_j$. 

The definition of the summary causal graph and full-time causal graph can be seen in the appendix.
\subsection{Related Work}
Traditionally, the Granger causality test involved comparing prediction errors from linear models with and without past information for two variables to assess their causal relationship through hypothesis testing.
However, this approach, based on linear Vector Autoregressive (VAR) had limitations in capturing all types of interactions. 
While some methods attempted to use kernel techniques to handle nonlinear systems \cite{Marinazzo_2008,amblard2012kernelizing}, they still lacked the flexibility of non-parametric models like transfer entropy and its invariants \cite{zhou2022causality}. 
Nonetheless, these non-parametric methods often require large amounts of data and suffer from slow computation, among other issues.
In recent years, machine learning, especially deep neural networks has been leveraged to enhance the capture of Granger causality, especially in dealing with nonlinear relationships and complex data.
As highlighted in \cite{tank2021neural}, integrating neural networks into the framework of Granger causality faces two key challenges: the lack of interpretability and the difficulty in disentangling the importance of variables in the presence of shared hidden layers for a joint network. This complicates the specification of selective Granger causality conditions.
To tackle these challenges, \cite{tank2021neural} chose to individually model each output component, apply a group sparse group lasso penalty on the weights of the first layer, and optimize by proximal gradient descent.

Most neural Granger Causality approaches follow the idea in \cite{tank2021neural}, utilizing various time series forecasting models to improve the interpretability of causal relations learned by black-box methods. For instance, \cite{khanna2020economy} employed an economy statistical recurrent units (eSRU) model, while \cite{bussmann2021neural} proposed an additive neural Vector Autoregressive (VAR) model. Additionally, \cite{suryadi2023granger} modified the architecture by transforming fully connected hidden layers from the input layer into a one-to-one configuration. Alternatively, \cite{Li_Yu_Principe_2023} utilizes a dynamic variational autoencoder model, which learns causality and enhances generative performance. Concerning the causal sparsity constraint, all mentioned methods enforce sparsity either on the weights of the first layer or the weights of the final layer.

Based on our current understanding, we observe that existing methods predominantly rely on single-variable time prediction models. This raises the following considerations:

\textit{\textbf{Q1}: Why do these methods primarily rely on single-variable prediction models? Why not utilize multivariate prediction models?}

\textbf{A1}: These methods depend on neural network weights, either in the first layer or the last layer, to express and constrain Granger causality. Utilizing single-variable neural network prediction models aids in decoupling the relationships between variables, whereas in multivariate prediction models, the shared hidden layer prevents the isolation of variable importance, as exemplified in Figure \ref{fig:model_figure}.

\textit{\textbf{Q2}: What are the existing issues with current methods?}

\textbf{A2}: These methods restrict the neural network's ability to learn complex data relationships and additional computation. 
In particular,  RNNs, which are suitable for non-stationary time series, fail to capture full-time Granger causality in the current framework.

\textit{\textbf{Q3}: How can we overcome the aforementioned challenges?}

\textbf{A3}: We propose constraining the Jacobian matrix between inputs and outputs. This approach eliminates the need to express Granger causality through network weights, enabling the learning of summary and full-time Granger causality on any multivariate time series prediction network that is suitable for the data characteristics.


\section{Method}
Given a $D$-dimensional multivariate time series $\mathbf{x}=\{x_1,x_2,\ldots.x_D\}$,  as shown in  Figure \ref{fig:model_figure}, our model can be summarized as follows:

\begin{enumerate}[label=\textbf{\arabic*.}]
	\item With the past information of each time series $\mathbf{x}_{1:D}^{t-\tau :t-1}$ as input,  we forecast their future $\mathbf{x}_{1:D}^{t}$ by a multivariate time series forecasting model.
 In this work, we leverage a residual MLP network.
	\item In the training phase, we train the model with mean squared error loss along with the input-output Jacobian matrix regularizer, which facilitates the model's learning of both Granger causality and forecasting.
	\item In the post-hoc analysis phase, we use the learned input-output Jacobian matrix to analyze the summary and full-time Granger causality between time series.
\end{enumerate}

In the following part, we will provide detailed descriptions of the proposed model, regularizer term, and post-hoc analysis method. 
Before that,  it is crucial to emphasize that we leverage the residual MLP network for its advantages in forecasting time series \cite{das2023longterm, Zeng_Chen_Zhang_Xu_2023}; however, our proposed framework is versatile and can be applied to alternative models, such as LSTM (see the Experiments section). With this approach, we can also achieve full-time Granger causality.

\subsection{Residual MLP-based time series forecasting model} 
\begin{figure}[htbp]
	\centering
	\includegraphics[width=0.4\textwidth]{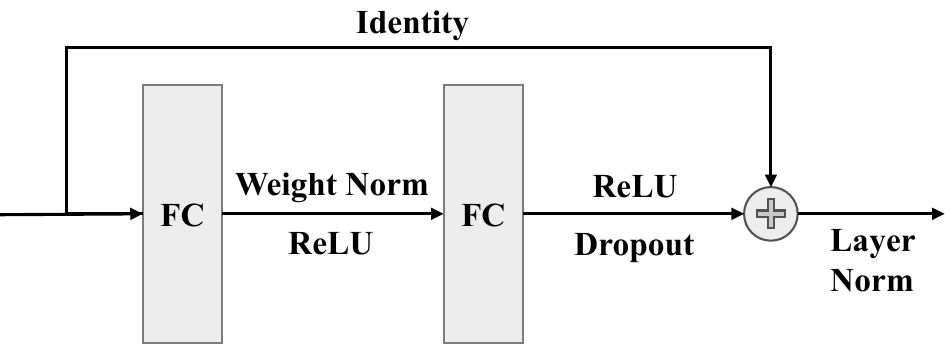}
	\caption{The framework of the Residual MLP layer in this work. FC represents the fully connected layer.}
	\label{fig:residualmlp}
\end{figure}

Granger causality relies on prediction as a fundamental component. 
It always uses the time series forecasting model to discover the Granger causality. 
A good predictive model with high performance, free from overfitting, and resilient to noise, can facilitate Granger causality learning.
Due to the success of residual MLP-based models in time series analysis \cite{das2023longterm, Zeng_Chen_Zhang_Xu_2023}, which are employed in state-of-the-art (SOTA) methods for predicting time series, we utilize a simple MLP-based model to learn Granger causality. 
Our model includes an input layer and an individual output layer, along with a varying number of robust residual MLP layers that seamlessly join together to create a unified framework for analyzing multivariate time series data, which can be formulated as follows:

\begin{equation}
	\mathbf{\hat{x}_{1:D}^{t}} = \mathrm{FC_2}(\mathrm{ResidualBlock}(\mathrm{FC_1}(\mathbf{x}_{1:D}^{t-\tau :t-1}))),
\end{equation}
where $\mathrm{FC_1}$ and $\mathrm{FC_2}$ represent fully-connected layers. $\mathrm{ResidualBlock}$ is composed of $n$ residual MLP layers, as shown in the Figure \ref{fig:residualmlp}.
Since this architecture does not have any self-attention, recurrent, or convolutional mechanisms, it allows our model to benefit from the simplicity and speed of linear models while still being able to handle non-linear dependencies.

\subsection{Input-output Jacobian regularizer as a Granger causality constraint}
It is necessary to incorporate the null hypothesis that the variable $x_i$ is not the Granger causality of $x_j$ as a Granger causality constraint within the model. Conventionally, this is accomplished by applying $L_1$ or $L_2$ norm regularizer to weights of the first layer \cite{tank2021neural, suryadi2023granger}. This regularizer approach contributes to learning the Granger causal connections between variables in the case of univariate time series prediction, where the weight can be expressed as a direct relationship from one variable to another.
However, this principle encounters three challenges. Firstly, imposing a penalty on the first layer may hinder the neural network from adequately learning the time series.
Secondly, the weights of the first layer prevent RNNs, LSTMs, and other recurrent networks from learning full-time Granger causality.
Thirdly, if one wishes to alter the current framework for learning Granger causality in multivariate multi-model settings, it becomes necessary to abandon this approach.
In contrast, the input-output Jacobian matrix allows us to find the relationships between variables,  including their interactions, self-importance, and dependencies over time.
The definition of the Jacobian matrix is as follows:
\begin{equation}
\begin{aligned}
\label{jac}
	\mathbf{J}&=\left[\begin{array}{lll}
		\frac{\partial \mathbf{f}}{\partial x_1^{t-\tau}}  \cdots  \frac{\partial \mathbf{f}}{\partial x_{D}^{t-1}}
	\end{array}\right]\\
        &=\left[\begin{array}{ccccccc}
		\frac{\partial f_1}{\partial x_{1}^{t-\tau}}  \cdots \frac{\partial f_1}{\partial x_{1}^{t-1}} \cdots \frac{\partial f_1}{\partial x_{D}^{t-\tau}} \cdots \frac{\partial f_1}{\partial x_{D}^{t-1}} \\
		\vdots  \\
		\frac{\partial f_{D}}{\partial x_{1}^{t-\tau}}  \cdots \frac{\partial f_{D}}{\partial x_{1}^{t-1}} \cdots \frac{\partial f_{D}}{\partial x_{D}^{t-\tau}} \cdots \frac{\partial f_{D}}{\partial x_{D}^{t-1}} \\
	\end{array}\right],
\end{aligned}
\end{equation}
where the $f_i$ represents the $i$-th predictive function for $i-$th variable. 

To detect the lags where Granger causal effects exist, we enforce the sparsity of the input-output Jacobian matrix.
Nevertheless, the complexity of computing the $L_1$ norm of a Jacobian matrix of size $(D, D\tau)$  is  $O(g(D, \tau))$,
and $g(D, \tau)$ represents the specific time and resource requirements of the algorithm.
It will cost lots of computing time when the $D$ is high, which can be seen in Table \ref{tab:time_param_var_100}.

Instead, we regularize the squared Frobienus norm of the input-output Jacobian matrix as \cite{hoffman2019robust}, which can be efficiently computed:
\begin{equation}
\begin{aligned}\label{JF1}
\|J(\boldsymbol{x})\|_{\mathrm{F}}^2 &= \operatorname{Tr}(J J^{\mathrm{T}})= \sum_{\{\boldsymbol{e}\}} \boldsymbol{e} J J^{\mathrm{T}} \boldsymbol{e}^{\mathrm{T}} \\
&= \sum_{\{\boldsymbol{e}\}}\left[\frac{\partial(\boldsymbol{e} \cdot \boldsymbol{z})}{\partial \boldsymbol{x}}\right]^2,
\end{aligned}
\end{equation}
where $\{e\}$ denotes a constant orthonormal basis of the $D$-dimensional output space.
$\boldsymbol{z}$ is the output with respect to input variables $\boldsymbol{x}$. Tr$(\cdot)$ represents the trace function.
Ultimately, this leads to computational overhead that increases linearly with the output dimension $D$.

As illustrated in \cite{hoffman2019robust}, we can rewrite the Eq.~\ref{JF1} and use random projection to compute the squared Frobienus norm of the input-output Jacobian matrix efficiently,  which projects the high-dimensional data onto a lower-dimensional space, thereby reducing computational and storage costs:
\begin{equation}
\|J(\boldsymbol{x})\|_{\mathrm{F}}^2 \approx \frac{1}{n_{\text {proj }}} \sum_{\mu=1}^{n_{\text {proj }}}\left[\frac{\partial\left(\hat{\boldsymbol{v}}^\mu \cdot \boldsymbol{z}\right)}{\partial \boldsymbol{x}}\right]^2,
\end{equation}
where the random vector $\hat{\boldsymbol{v}}^\mu$ is drawn from the $(D-1)$-dimensional unit sphere $S^{D-1}$, $n_{\text {proj }}$ is the number of random projection. Utilizing a mini-batch size of $|B|$ = 100, a singular projection results in model performance nearly indistinguishable from the exact method, while significantly reducing computational costs by orders of magnitude.

Above all, our method can be named JRNGC-L1, JRNGC-F according to different norm regularizers.
The penalized loss function of JRNGC-L1 can be formulated as follows:
\begin{equation}
\label{eq:L1}
\frac{1}{D} \sum_{i=1}^{D} \frac{1}{N - \tau}\sum_{t=\tau +1}^{N} (\hat{x}_{i,t} - x_{i,t})^2 + \lambda \|J(\boldsymbol{x})\|_1.
\end{equation}

The penalized loss function of JRNGC-F can be formulated as follows:
\begin{equation}
\label{eq:F}
\frac{1}{D} \sum_{i=1}^{D} \frac{1}{N - \tau} \sum_{t=\tau +1}^{N} (\hat{x}_{i,t} - x_{i,t})^2 + \lambda \|J(\boldsymbol{x})\|_{\mathrm{F}}^2,
\end{equation}
where, in both Eq. \ref{eq:L1} and Eq. \ref{eq:F}, the $\lambda$ mean the regularizer coefficient and $D$ is the dimension of time series and $N$ is the length of time.
Note that, although the Jacobian regularizer is popular in computer vision \cite{jakubovitz2018improving,rhodes2021local}, to our best knowledge, we are the first to use it in Granger causality learning.

\subsection{Input-output Jacobian matrix as the variable's causal importance}
In the post-hoc analysis, we employ the learned input-output Jacobian matrix to analyze the causal relationships between variables.
The input-output Jacobian matrix can obtain the variable's causal importance through $J_{i,j,\alpha} = \frac{\partial x_{j}}{\partial x_{i}^{(t-\alpha)}} $.
It allows us to obtain the variable's lag importance, i.e., the full-time Granger causal graph.

\begin{table*}[ht]
\small
\caption{Comparisons of AUROC and AUPRC for Granger causality among different approaches on VAR dataset. The mean results with standard error are reported by averaging over 5 runs. Note that \enquote{with lag} refers to the full-time causal graph, and \enquote{with no lag} refers to the summary causal graph.}
\vskip 0.15in
	\centering 
		\begin{tabular}{lccccccc}
			\toprule
			\multicolumn{1}{c}{\multirow{2}{*}{Type}} & Model  & \multicolumn{2}{c}{VAR$(100, 5, 10)$} & \multicolumn{2}{c}{VAR$(50, 5, 10)$} & \multicolumn{2}{c}{VAR$(10, 3, 5)$} \\
			\cmidrule(r){2-2} \cmidrule(r){3-4} \cmidrule(r){5-6} \cmidrule(r){7-8}
			\multicolumn{1}{c}{} & Metric & AUROC ($\uparrow$) & AUPRC ($\uparrow$) & AUROC ($\uparrow$) & AUPRC ($\uparrow$) & AUROC ($\uparrow$) & AUPRC ($\uparrow$) \\
			\midrule
			\multirow{4}{*}{with lag} 
			& cMLP & 0.973\smallgrey{$\pm$0.013} & 0.770\smallgrey{$\pm$0.131} & 0.995\smallgrey{$\pm$0.005} & 0.893\smallgrey{$\pm$0.061} & 0.994\smallgrey{$\pm$0.009} & 0.912\smallgrey{$\pm$0.127}  \\ 
			& JGC & \textbf{1.000}\smallgrey{$\pm$0.000} & \textbf{1.000}\smallgrey{$\pm$0.000} & \textbf{1.000}\smallgrey{$\pm$0.000} & \textbf{1.000}\smallgrey{$\pm$0.000} & \textbf{1.000}\smallgrey{$\pm$0.000} & \textbf{0.998}\smallgrey{$\pm$0.003} \\
			& \textbf{JRNGC-L1(ours)} & \textbf{1.000}\smallgrey{$\pm$0.000} & \textbf{1.000}\smallgrey{$\pm$0.000} & \textbf{1.000}\smallgrey{$\pm$0.000} & \textbf{1.000}\smallgrey{$\pm$0.000} & \textbf{1.000}\smallgrey{$\pm$0.000} & \textbf{0.998}\smallgrey{$\pm$0.004}  \\ 
			& \textbf{JRNGC-F(ours)} & \underline{0.997}\smallgrey{$\pm$0.001} & \underline{0.950}\smallgrey{$\pm$0.006} & \textbf{1.000}\smallgrey{$\pm$0.001} & \underline{0.994}\smallgrey{$\pm$0.002} & 0.999\smallgrey{$\pm$0.002} & \underline{0.993}\smallgrey{$\pm$0.013} \\ 
			
			\midrule
			\multirow{8}{*}{with no lag} 
			& cMLP  & 0.940\smallgrey{$\pm$0.0013} & 0.851\smallgrey{$\pm$0.051} & 0.973\smallgrey{$\pm$0.022} & 0.929\smallgrey{$\pm$0.045} & 0.978\smallgrey{$\pm$0.032} & 0.923\smallgrey{$\pm$0.110}\\
			& cLSTM  & 0.845\smallgrey{$\pm$0.045} & 0.606\smallgrey{$\pm$0.102} & 0.935\smallgrey{$\pm$0.018} & 0.816\smallgrey{$\pm$0.051} & 0.931\smallgrey{$\pm$0.061} & 0.803\smallgrey{$\pm$0.162} \\
			& NAVAR(MLP)  & 0.887\smallgrey{$\pm$0.023} & 0.776\smallgrey{$\pm$0.040} &  0.960\smallgrey{$\pm$0.019} & 0.909\smallgrey{$\pm$0.043} & 0.976\smallgrey{$\pm$0.131} & 0.936\smallgrey{$\pm$0.086}  \\
			& NAVAR(LSTM)  & 0.860\smallgrey{$\pm$0.011} & 0.750\smallgrey{$\pm$0.018} & 0.930\smallgrey{$\pm$0.016} & 0.851\smallgrey{$\pm$0.022} & 0.952\smallgrey{$\pm$0.064} & 0.875\smallgrey{$\pm$0.156} \\
			& JGC & \textbf{1.000}\smallgrey{$\pm$0.000} & \textbf{1.000}\smallgrey{$\pm$0.000}& \textbf{1.000}\smallgrey{$\pm$0.000} & \textbf{1.000}\smallgrey{$\pm$0.000} & \textbf{1.000}\smallgrey{$\pm$0.001} & \underline{0.998}\smallgrey{$\pm$0.003}  \\
			& CR-VAE& 0.645\smallgrey{$\pm$0.014} &0.268\smallgrey{$\pm$0.008} & 0.638\smallgrey{$\pm$0.014} &0.282\smallgrey{$\pm$0.017}&0.749\smallgrey{$\pm$0.013}&0.401\smallgrey{$\pm$0.107}\\
			& \textbf{JRNGC-L1(ours)}  & \textbf{1.000}\smallgrey{$\pm$0.000} & \textbf{1.000}\smallgrey{$\pm$0.000} & \textbf{1.000}\smallgrey{$\pm$0.000} & \textbf{1.000}\smallgrey{$\pm$0.000} & \textbf{1.000}\smallgrey{$\pm$0.001} & \textbf{0.999}\smallgrey{$\pm$0.003}  \\
			& \textbf{JRNGC-F(ours)}  & \underline{0.984}\smallgrey{$\pm$0.002} & \underline{0.960}\smallgrey{$\pm$0.005} & \underline{0.998}\smallgrey{$\pm$0.001} & \underline{0.995}\smallgrey{$\pm$0.001} & 0.996\smallgrey{$\pm$0.007} & 0.994\smallgrey{$\pm$0.011} \\
			
			\bottomrule
		\end{tabular}
 \label{tab:var}
 \vskip -0.1in
\end{table*}

\section{Experiments}

In this section, we demonstrate the performance of the proposed methods, i.e., JRNGC-L1, JRNGC-F on five widely used benchmarks: the VAR model, the Lorenz-96 model, fMRI data, the DREAM-3 dataset and CausalTime \cite{cheng2024causaltime}. 
We perform comparative experiments with competitive methods, including GC \cite{granger1969investigating}, PCMCI \cite{runge2019detecting}, cMLP \cite{tank2021neural}, cLSTM \cite{tank2021neural}, NAVAR (MLP) \cite{bussmann2021neural}, NAVAR (LSTM) \cite{bussmann2021neural},  SRU, eSRU\cite{khanna2020economy}, TCDF \cite{nauta2019causal}, JGC \cite{suryadi2023granger} and CR-VAE \cite{Li_Yu_Principe_2023}, Scalable Causal Graph Learning (SCGL, \cite{xu2019scalable}), CUTS \cite{yuxiao2023cuts}, CUTS+ \cite{cheng2024cuts+}, NTS-NOTEARS (abbreviated as N.NTS, \cite{sun2023nts}), Rhino \cite{gong2023rhino}, and Latent Convergent Cross Mapping (LCCM, \cite{brouwer2021latent}), as well as Neural Graphical Model (NGM, \cite{bellot2022neural}).
Additionally, our summary hyperparameters of experiments for all models and other additional experiments are detailed in the appendix.

\subsection{Metrics}

We employ two standard metrics: the area under the receiver operating characteristic curve (AUROC) and the area under the precision-recall curve (AUPRC). A value of 0.5 or lower in AUROC indicates poor performance. In scenarios with sparse causal relationships, AUPRC becomes a more reliable indicator of the model's ability to detect causal edges. This is due to its emphasis on correctly identifying positive instances, a critical aspect when the number of actual causal relationships is limited.

\subsection{Experiment results and analysis}

In this section, we demonstrate the effectiveness of our proposed method in inferring Granger causality between variables, i.e., the summary Granger causality, and the full-time Granger causality.
Specifically, we use the term \enquote{with no lag} to express Granger causality between variables, and \enquote{with lag} to express full-time Granger causality. 
Our results show that our proposed method is capable of learning summary Granger causality and full-time Granger causality with high AUROC and AUPRC scores on both synthetic and open-real benchmark datasets.

\textbf{VAR model.} For this model, we simulated up to $T=600$ observations with the maximum true time lag $\tau\in \{3, 5\}$, $D \in \{10, 50, 100\}$, maximum estimated lag $\eta \in \{5, 10\}$ and $\tau \sim N(0, I)$ to demonstrate the ability to deal with high-dimensional time series.
We conduct experiments on five independently generated realizations of the system.
From Table \ref{tab:var}, we observe that using an input-output Jacobian matrix in the post-hoc analysis yields a more comprehensive and informative grasp of the Granger causal relationships within the data when compared to other analogous techniques like cMLP, cLSTM, NAVAR(MLP), NAVAR(LSTM), and CR-VAE.
Although JGC also benefits from using the input-output Jacobian matrix to estimate Granger causality, 
our methods JRNGC-F and JRNGC-L1 offer distinct advantages. 
They effectively integrate the Jacobian matrix into both training and post-hoc phases, resulting in remarkable performance, especially when dealing with more complex datasets than the VAR dataset, such as the Lorenz dataset.

\textbf{Lorenz-96.} For this dataset, we simulated up to $T=500$ observations and we set $D=10, F\in \{10,40\}$. $F$ is a forcing constant, and higher values of $F$ can lead to increased chaotic behavior, enhanced nonlinearity \cite{Karimi_2010}.
When $F$ is 40, the time series becomes more challenging to predict than when $F$ is 10.
We conduct experiments on five independently generated realizations of the system.
As can be seen from Table \ref{tab:lor}, our methods JRNGC-F and JRNGC-L1 achieve the best scores and robust AUPRC scores, especially when the $F=40$, i.e., the relationships between variables are more complex.

Both the VAR datasets and Lorenz-96 dataset possess true summary Granger causality and full-time Granger causality. From Table \ref{tab:var} and \ref{tab:lor}, we observe our methods consistently exhibit robust performance in estimating the full-time Granger causality (denoted as the \enquote{with lag} performance) and the summary Granger causality (denoted as the \enquote{with no lag} performance). 
In contrast, other methods either fail to capture full-time causal relationships (e.g., the cLSTM) or experience evident performance degradation when attempting to learn full-time causal relationships (e.g., cMLP, JGC for Lorenz-96 dataset with F = $40$).
The datasets introduced below, namely fRMI, DREAM-3 datasets and CausalTime, exclusively feature true summary causal graphs.

\textbf{fMRI data.} Here, we select data from the first subject in the third set of simulations for the experiment.
We ran five independent experiments on the selected subject.
Table \ref{tab:fmri} summarizes the performance of each method on fMRI data.
Importantly, the true variable usage rate in this experiment is only 14.67\%, emphasizing the sparsity of the data.
In such data, identifying positive connections between different brain regions becomes crucial.  Hence, we need to pay close attention to the AUPRC score, which reflects our method's ability to capture these sparse relationships. Remarkably, our methods demonstrate the capability to achieve a competitive AUROC score while simultaneously outperforming in terms of AUPRC, making them well-suited for this challenging context.

\begin{table*}[htbp]
\small
\caption{Comparisons of AUROC and AUPRC for Granger causality among different approaches on Lorenz-96 dataset. The mean results with standard error are reported by averaging over 5 runs. Note that \enquote{with lag} refers to the full-time causal graph, and \enquote{with no lag} refers to the summary causal graph.}
\vskip 0.15in
	\centering 
	\begin{tabular}{lccccccc}
		\toprule
		\multicolumn{1}{c}{\multirow{2}{*}{Type}} & Model & \multicolumn{2}{c}{F=10} & \multicolumn{2}{c}{F=40}  \\
		\cmidrule(r){2-2} \cmidrule(r){3-4} \cmidrule(r){5-6}
		\multicolumn{1}{c}{} & Metric & AUROC ($\uparrow$) & AUPRC ($\uparrow$) & AUROC ($\uparrow$) & AUPRC ($\uparrow$) \\
		\midrule
		\multirow{5}{*}{with lag} 
		& cMLP & \underline{0.999}\smallgrey{$\pm$0.002} & \underline{0.992}\smallgrey{$\pm$0.016} & 0.998\smallgrey{$\pm$0.001} & 0.968\smallgrey{$\pm$0.016}  \\ 
		& JGC & 0.998\smallgrey{$\pm$0.001} & 0.979\smallgrey{$\pm$0.011} & \underline{0.992}\smallgrey{$\pm$0.006} & 0.939\smallgrey{$\pm$0.020}   \\ 
		& \textbf{JRNGC-L1(ours)} & \textbf{1.000}\smallgrey{$\pm$0.000} & \textbf{1.000}\smallgrey{$\pm$0.000} & \textbf{0.999}\smallgrey{$\pm$0.000} & \textbf{0.994}\smallgrey{$\pm$0.003}  \\
		& \textbf{JRNGC-F(ours)} & \textbf{1.000}\smallgrey{$\pm$0.000} & \textbf{1.000}\smallgrey{$\pm$0.000} & \textbf{0.999}\smallgrey{$\pm$0.001} & \underline{0.989}\smallgrey{$\pm$0.009}  \\
		\midrule
		\multirow{7}{*}{with no lag} 
		& cMLP & \underline{0.997}\smallgrey{$\pm$0.006} & \underline{0.998}\smallgrey{$\pm$0.005} & 0.984\smallgrey{$\pm$0.007} & 0.968\smallgrey{$\pm$0.016}  \\ 
		& cLSTM & 0.974\smallgrey{$\pm$0.028} & 0.949\smallgrey{$\pm$0.068} & 0.896\smallgrey{$\pm$0.007} & 0.854\smallgrey{$\pm$0.008}  \\ 
		& NAVAR(MLP) & 0.993\smallgrey{$\pm$0.004} & 0.986\smallgrey{$\pm$0.008} & 0.900\smallgrey{$\pm$0.021} &0.828\smallgrey{$\pm$0.052}   \\ 
		& NAVAR(LSTM) & 0.993\smallgrey{$\pm$0.006} & 0.988\smallgrey{$\pm$0.011} & 0.891\smallgrey{$\pm$0.036} & 0.810\smallgrey{$\pm$0.056}  \\ 
		& JGC & 0.989\smallgrey{$\pm$0.006} & 0.982\smallgrey{$\pm$0.010} & 0.956\smallgrey{$\pm$0.024} & 0.943\smallgrey{$\pm$0.016}   \\ 
		&CR-VAE &0.898\smallgrey{$\pm$0.020} &0.776\smallgrey{$\pm$0.041} &0.654\smallgrey{$\pm$0.030}&0.493\smallgrey{$\pm$0.023}\\
		& \textbf{JRNGC-L1(ours)} & \textbf{1.000}\smallgrey{$\pm$0.000} & \textbf{1.000}\smallgrey{$\pm$0.000} & \textbf{0.995}\smallgrey{$\pm$0.002} & \textbf{0.994}\smallgrey{$\pm$0.003}  \\ 
		& \textbf{JRNGC-F(ours)} & \textbf{1.000}\smallgrey{$\pm$0.000} & \textbf{1.000}\smallgrey{$\pm$0.000} & \underline{0.992}\smallgrey{$\pm$0.006} & \underline{0.989}\smallgrey{$\pm$0.009}  \\ 
		\bottomrule
	\end{tabular}
\label{tab:lor}
\vskip -0.1in
\end{table*}

\begin{table}[ht]
\small
 \caption{Comparative performance of the summary causal graph for the first subject in fMRI. The mean results with standard error are reported by averaging over $5$ runs. }
 \vskip 0.15in
	\centering
	\begin{tabular}{ccc}
		\toprule
		Model & AUROC ($\uparrow$) & AUPRC ($\uparrow$)  \\ 
		\midrule
		cMLP & 0.875\smallgrey{$\pm$0.014} & \underline{0.726}\smallgrey{$\pm$0.013} \\ 
		cLSTM & 0.845\smallgrey{$\pm$0.029} & 0.644\smallgrey{$\pm$0.079} \\ 
		NAVAR(MLP) & 0.840\smallgrey{$\pm$0.004} & 0.477\smallgrey{$\pm$0.014} \\ 
		NAVAR(LSTM) & 0.742\smallgrey{$\pm$0.015} & 0.320\smallgrey{$\pm$0.020} \\ 
		eSRU & 0.743\smallgrey{$\pm$0.047} & 0.514\smallgrey{$\pm$0.057} \\
		CR-VAE & 0.516\smallgrey{$\pm$0.029} & 0.152\smallgrey{$\pm$0.007} \\
		JGC & \textbf{0.912}\smallgrey{$\pm$0.007} & 0.536\smallgrey{$\pm$0.009}  \\  
		\textbf{JRNGC-L1(ours)} & 0.845\smallgrey{$\pm$0.011} & 0.707\smallgrey{$\pm$0.004} \\ 
		\textbf{JRNGC-F(ours)} & \underline{0.898}\smallgrey{$\pm$0.001} & \textbf{0.749}\smallgrey{$\pm$0.003} \\
		\bottomrule
	\end{tabular}
\label{tab:fmri}
\vskip -0.1in
\end{table}

\begin{table}[ht]
\caption{AUROC scores of summary causal graph for the five datasets in DREAM-3, alongside corresponding true variable usage percentages. 
 Here, E1 refers to E.coli-1, Y1 refers to Yeast-1, and so on.}
 \vskip 0.15in
	\centering
	\resizebox{0.472 \textwidth}{!}{
		\begin{tabular}{cccccc}
			\toprule
			Model & E1 & E2 & Y1 & Y2 & Y3  \\ 
			\midrule
			cMLP$^{*}$ & 0.644 & 0.568 & 0.585 & 0.506 & 0.528  \\ 
			cLSTM$^{*}$& 0.629 & 0.609 & 0.579 & 0.519 & 0.555  \\ 
			TCDF$^{*}$ & 0.614 & 0.647 & 0.581 & 0.556 & \underline{0.557}  \\ 
			SRU$^{*}$ & 0.657 & \underline{0.666} & 0.617 & 0.575 & 0.550  \\ 
			eSRU$^{*}$ & \underline{0.660} & 0.629 & 0.627 & 0.557 & 0.550 \\ 
			NAVAR(MLP) & 0.557 & 0.577 & \textbf{0.652} & \underline{0.573} & 0.548  \\ 
			NAVAR(LSTM) & 0.544 & 0.473 & 0.497 & 0.477 & 0.466 \\
			JGC & 0.504 &0.527 &0.604 &0.553 &0.521\\
			CR-VAE &0.502&0.494&0.525&0.518&0.501\\
			\textbf{JRNGC-F(ours)} & \textbf{0.666} & \textbf{0.678} & \underline{0.650} & \textbf{0.597} & \textbf{0.560}  \\ 
			\midrule
			True variable usage & 1.25\% & 1.19\% & 1.66\% & 3.89\% & 5.51\% \\
			\bottomrule
		\end{tabular}
	}
\label{tab:dream}
\vskip -0.1in
\end{table}

\begin{table*}[ht]
\small
\centering
\caption{Comparative performance on CausalTime benchmark datasets. We
highlight the best and the second best in bold and with underlining, respectively.}
\vskip 0.15in
\label{tab:ra}
\begin{tabular}{@{}lcccccc@{}}
\toprule
Methods & \multicolumn{3}{c}{AUROC} & \multicolumn{3}{c}{AUPRC} \\
\cmidrule(lr){2-4} \cmidrule(lr){5-7} & AQI & Traffic & Medical & AQI & Traffic & Medical \\
\midrule
GC     & $0.4538 \smallgrey{$\pm$ 0.0377}$ & $0.4191 \smallgrey{$\pm$ 0.0310}$ & $0.5737 \smallgrey{$\pm$ 0.0338}$ & $0.6347 \smallgrey{$\pm$ 0.0158}$ & $0.2789 \smallgrey{$\pm$ 0.0018}$ & $0.4213 \smallgrey{$\pm$ 0.0281}$ \\
SVAR   & $0.6225 \smallgrey{$\pm$ 0.0406}$ & $0.6329 \smallgrey{$\pm$ 0.0047}$ & $0.7130 \smallgrey{$\pm$ 0.0188}$ & $0.7903 \smallgrey{$\pm$ 0.0175}$ & $0.5845 \smallgrey{$\pm$ 0.0021}$ & $0.6774 \smallgrey{$\pm$ 0.0358}$ \\
N.NTS  & $0.5729 \smallgrey{$\pm$ 0.0229}$ & \underline{$0.6329 \smallgrey{$\pm$ 0.0335}$} & $0.5019 \smallgrey{$\pm$ 0.0682}$ & $0.7100 \smallgrey{$\pm$ 0.0228}$ & $0.5770 \smallgrey{$\pm$ 0.0542}$ & $0.4567 \smallgrey{$\pm$ 0.0162}$ \\
PCMCI  & $0.5272 \smallgrey{$\pm$ 0.0744}$ & $0.5422 \smallgrey{$\pm$ 0.0737}$ & $0.6991 \smallgrey{$\pm$ 0.0111}$ & $0.6734 \smallgrey{$\pm$ 0.0372}$ & $0.3474 \smallgrey{$\pm$ 0.0581}$ & $0.5082 \smallgrey{$\pm$ 0.0177}$ \\
Rhino  & $0.6700 \smallgrey{$\pm$ 0.0983}$ & $0.6274 \smallgrey{$\pm$ 0.0185}$ & $0.6520 \smallgrey{$\pm$ 0.0212}$ & $0.7593 \smallgrey{$\pm$ 0.0755}$ & $0.3772 \smallgrey{$\pm$ 0.0093}$ & $0.4897 \smallgrey{$\pm$ 0.0321}$ \\
CUTS & $0.6013 \smallgrey{$\pm$0.0038}$ & $0.6238 \smallgrey{$\pm$0.0179}$ & $0.3739 \smallgrey{$\pm$0.0297}$ & $0.5096 \smallgrey{$\pm$0.0362}$ & $0.1525 \smallgrey{$\pm$0.0226}$ & $0.1537 \smallgrey{$\pm$0.0039}$ \\
CUTS+ & \underline{$0.8928 \smallgrey{$\pm$0.0213}$} & $0.6175 \smallgrey{$\pm$0.0752}$ & \textbf{0.8202 \smallgrey{$\pm$0.0173}} & \underline{$0.7983 \smallgrey{$\pm$0.0875}$} & \textbf{0.6367 \smallgrey{$\pm$0.1197}} & $0.5481 \smallgrey{$\pm$0.1349}$ \\
NGC & $0.7172 \smallgrey{$\pm$0.0076}$ & $0.6032 \smallgrey{$\pm$0.0056}$ & $0.5744 \smallgrey{$\pm$0.0096}$ & $0.7177 \smallgrey{$\pm$0.0069}$ & $0.3583 \smallgrey{$\pm$0.0495}$ & $0.4637 \smallgrey{$\pm$0.0121}$ \\
NGM & $0.6728 \smallgrey{$\pm$0.0164}$ & $0.4660 \smallgrey{$\pm$0.0144}$ & $0.5551 \smallgrey{$\pm$0.0154}$ & $0.4786 \smallgrey{$\pm$0.0196}$ & $0.2826 \smallgrey{$\pm$0.0098}$ & $0.4697 \smallgrey{$\pm$0.0166}$ \\
LCCM & $0.8565 \smallgrey{$\pm$0.0653}$ & $0.5545 \smallgrey{$\pm$0.0254}$ & \underline{$0.8013 \smallgrey{$\pm$0.0218}$} & \textbf{0.9260 \smallgrey{$\pm$0.0246}} & $0.5907 \smallgrey{$\pm$0.0475}$ & \textbf{0.7554 \smallgrey{$\pm$0.0235}} \\
eSRU & $0.8229 \smallgrey{$\pm$0.0317}$ & $0.5987 \smallgrey{$\pm$0.0192}$ & $0.7559 \smallgrey{$\pm$0.0365}$ & $0.7223 \smallgrey{$\pm$0.0317}$ & $0.4886 \smallgrey{$\pm$0.0338}$ & \underline{$0.7352 \smallgrey{$\pm$0.0600}$} \\
SCGL & $0.4915 \smallgrey{$\pm$0.0476}$ & $0.5927 \smallgrey{$\pm$0.0553}$ & $0.5019 \smallgrey{$\pm$0.0224}$ & $0.3584 \smallgrey{$\pm$0.0281}$ & $0.4544 \smallgrey{$\pm$0.0315}$ & $0.4833 \smallgrey{$\pm$0.0185}$ \\
TCDF & $0.4148 \smallgrey{$\pm$0.0207}$ & $0.5029 \smallgrey{$\pm$0.0041}$ & $0.6329 \smallgrey{$\pm$0.0384}$ & $0.6527 \smallgrey{$\pm$0.0087}$ & $0.3637 \smallgrey{$\pm$0.0048}$ & $0.5544 \smallgrey{$\pm$0.0313}$ \\
\textbf{JRNGC-F (ours)} & \textbf{0.9279 \smallgrey{$\pm$0.0011}} & \textbf{0.7294 \smallgrey{$\pm$0.0046}} & $0.7540 \smallgrey{$\pm$0.0040}$ & $0.7828 \smallgrey{$\pm$0.0020}$ & \underline{$0.5940 \smallgrey{$\pm$0.0067}$} & $0.7261 \smallgrey{$\pm$0.0016}$ \\

\bottomrule
\end{tabular}
\vskip -0.1in
\end{table*}

\begin{table*}[ht]
\small
\caption{Evaluation of the impact of  Residual MLP Layer Depth and Jacobian regularizer. JR-F denotes F-norm Jacobian regularizer.}
\vskip 0.15in
	\centering
		\begin{tabular}{cccccccc}
			\toprule
			\multirow{2}{*}{JR-F} & \multirow{2}{*}{Residual Layer} & \multicolumn{2}{c}{VAR100} & \multicolumn{2}{c}{fMRI} & \multicolumn{2}{c}{Lorenz-96} \\
			\cmidrule(r){3-8}
			&  & AUROC & AUPRC & AUROC & AUPRC & AUROC & AUPRC \\
			\midrule
			$\times$ & 0 & 0.969\smallgrey{$\pm$0.003} & 0.667\smallgrey{$\pm$0.023} & 0.764\smallgrey{$\pm$0.000} & 0.266\smallgrey{$\pm$0.000} & 0.813\smallgrey{$\pm$0.018} & 0.571\smallgrey{$\pm$0.033} \\
			$\checkmark$ & 0 & \textbf{0.997}\smallgrey{$\pm$0.001} & \textbf{0.950}\smallgrey{$\pm$0.006} & \textbf{0.898}\smallgrey{$\pm$0.001} & \textbf{0.749}\smallgrey{$\pm$0.003} & 0.875\smallgrey{$\pm$0.010} & 0.692\smallgrey{$\pm$0.036} \\
			$\times$ & 1 & 0.936\smallgrey{$\pm$0.009} & 0.501\smallgrey{$\pm$0.032} & 0.729\smallgrey{$\pm$0.027} & 0.175\smallgrey{$\pm$0.023} & 0.954\smallgrey{$\pm$0.021} & 0.772\smallgrey{$\pm$0.061} \\
			$\checkmark$ & 1 & 0.993\smallgrey{$\pm$0.002} & 0.894\smallgrey{$\pm$0.014} & 0.836\smallgrey{$\pm$0.018} & 0.448\smallgrey{$\pm$0.013} & 0.995\smallgrey{$\pm$0.003} & 0.959\smallgrey{$\pm$0.021} \\
			$\times$ & 5 & 0.957\smallgrey{$\pm$0.005} & 0.606\smallgrey{$\pm$0.032} & 0.766\smallgrey{$\pm$0.026} & 0.227\smallgrey{$\pm$0.015} & 0.995\smallgrey{$\pm$0.004} & 0.959\smallgrey{$\pm$0.025} \\
			$\checkmark$ & 5 & 0.966\smallgrey{$\pm$0.008} & 0.688\smallgrey{$\pm$0.047} & 0.748\smallgrey{$\pm$0.032} & 0.329\smallgrey{$\pm$0.013} & \textbf{0.999}\smallgrey{$\pm$0.001} & \textbf{0.989}\smallgrey{$\pm$0.009} \\
			\bottomrule
		\end{tabular}
 
	\label{tab:ablation}
 \vskip -0.1in
\end{table*}

\begin{table*}[ht]
\small
\centering
\caption{Ablation study comparing the efficacy of different regularizers and ground-truth graph structures in detecting summary causal graphs. {JR-DAG} denotes the DAG regularizer, and {JR-F} denotes the F-norm regularizer, both applied to the input-output Jacobian matrix.}
\vskip 0.15in
\label{tab:ablation-study}
\begin{tabular}{lccccc}
\toprule
Dataset & Ground-truth Structure & Regularizer & AUROC & AUPRC  \\
\midrule
VAR (100, 5, 10) & DAG  & JR-DAG & 0.968 & 0.922   \\
VAR (100, 5, 10) & DAG  & JR-F & 0.984 & 0.959 \\
DREAM4 & have-cycles & JR-DAG & 0.614 & 0.042  \\
DREAM4 & have-cycles & JR-F & 0.766 & 0.253  \\
\bottomrule
\end{tabular}
\vskip -0.1in
\end{table*}

\begin{table}[htbp]
\small
\caption{AUROC and AUPRC performances on the Lorenz dataset ($F=40$) when adopting the Jacobian regularizer to the LSTM.}
\vskip 0.15in
\centering
\begin{tabular}{lccc}
\toprule
Type & Model & AUROC ($\uparrow$) & AUPRC ($\uparrow$) \\
\midrule
\multirow{2}{*}{with lag} &
LSTM & $0.956 \pm 0.005$ & $0.538 \pm 0.045$ \\
& +JacoR & $\textbf{0.992} \pm 0.003$ & $\textbf{0.929} \pm 0.014$ \\
\midrule
\multirow{3}{*}{with no lag} &
LSTM & $0.661 \pm 0.040$ & $0.539 \pm 0.045$ \\
& +JacoR & $\textbf{0.942} \pm 0.019$ & $\textbf{0.929} \pm 0.014$ \\
& cLSTM & $0.896 \pm 0.007$ & $0.854 \pm 0.008$ \\
\bottomrule
\end{tabular}
\label{tab:anotherar}
\vskip -0.1in
\end{table}

\begin{table}[htbp]
\small
 \caption{Overfitting Analysis of JRNGC-F and same model without Jacobian regularizer (w/o JR) on Lorenz-96 dataset with $F=40$.}
 \vskip 0.15in
	\centering
	\begin{tabular}{cccc}
		\toprule
		Model & MSE  & training loss & testing loss \\ 
		\midrule
		JRNGC & 0.878 & 0.0197 & 0.0830 \\ 
		w/o JR & 1.269 & 0.0047 & 0.1088\\
		\bottomrule
	\end{tabular}
\label{tab:lorenzf40mse1}
\vskip -0.1in
\end{table}

\textbf{DREAM-3 dataset.}
We followed the settings in the baseline method eSRU, and the experimental results for the algorithms marked with an asterisk ($*$) were directly referenced from the SRU method \cite{khanna2020economy}.
From Table \ref{tab:dream}, we can observe that all the methods can not achieve good performance on the DREAM-3 dataset.
This is because there are too few observations in the DREAM3 dataset, which makes the neural network method overfit the data.
However, an important observation can be made regarding the minimal usage of true variables, highlighting the sparsity of the causality existed in the dataset. 
Subsequently, when referring to the results in Table \ref{tab:dreamappen} in the appendix, we can observe that our method not only achieves a comparable AUROC score but also maintains a competitive AUPRC score. This consistency across both metrics demonstrates the robustness of our approach in accurately identifying the sparse Granger causality relationships within the dataset.

\textbf{CausalTime.} For the new benchmark datasets, we followed the experimental settings described by \cite{cheng2024causaltime}. 
Aside from our algorithm, \textbf{JRNGC}, all additional results presented in Table \ref{tab:ra} were directly reproduced from the CausalTime \cite{cheng2024causaltime}, which benchmarked the performance of several recent and representative causal discovery methods.
As illustrated in Table \ref{tab:ra}, among the methods evaluated, JRNGC-F, alongside CUTS+ and LCCM, consistently emerged as top performers, and most of the baselines do not get AUROC $> 0.9$.
In contrast, our proposed algorithm, JRNGC-F, demonstrates novel SOTA performance on the AQI and Traffic datasets, as evidenced by its AUROC metrics, achieving scores of AUROC $0.9279 \pm 0.0011$ and AUROC $0.7294 \pm 0.0046$ respectively. 
In addition, JRNGC-F demonstrates its reliability and robustness across all datasets, consistently achieving metrics well above the baseline threshold of $0.5$.

\begin{figure}[ht]
	\centering
	\includegraphics[width=0.47\textwidth]{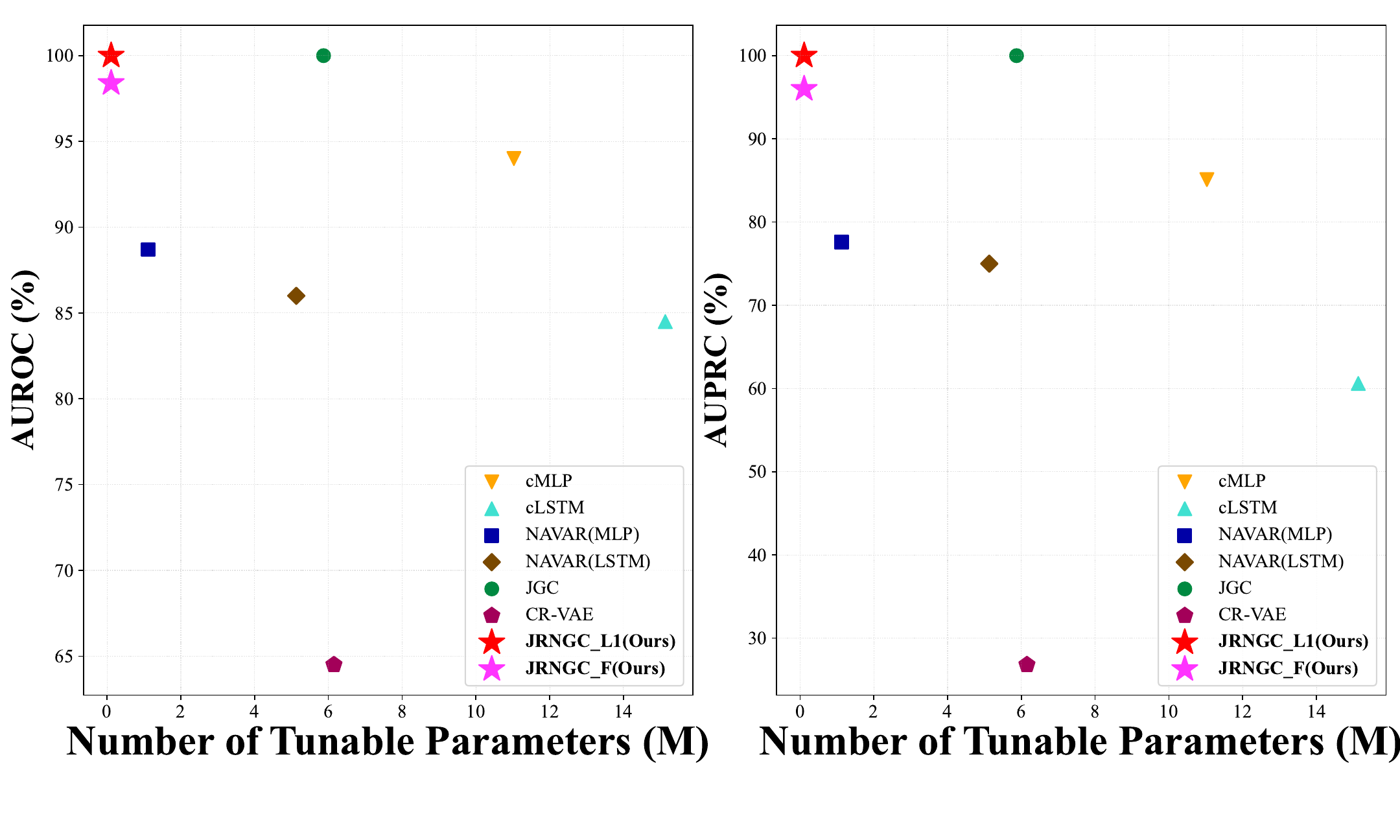}
	\caption{Performance comparisons on a 100-dimensional VAR dataset: AUROC, AUPRC, and the number of tunable parameters. }

	\label{fig:var100param}
\end{figure}

\textbf{Model complexity.}
The motivation behind proposing a unified neural Granger causality learning framework arises from the need for a single model to handle forecasting. However, the current neural Granger causality framework requires duplicating models for each variable to learn Granger causality, which is not only inelegant but also comes with drawbacks, as mentioned earlier.
Overcoming the challenge of learning Granger causality with multiple models could lead to a substantial reduction in model parameters to one divided by D (this could be achieved by utilizing the same neural network or a simpler yet more effective neural network.), where D denotes the dimensionality of the variables. 
As illustrated in Figure \ref{fig:var100param}, our proposed method, JRNGC, surpasses existing popular neural Granger causality methods with significantly lower model complexity and higher accuracy. 
For additional analyses, please refer to the appendix.

\subsection{Ablation studies}
In this section, we conduct ablation studies to investigate the individual contributions of various components of our methodology to the overall performance. These experiments are designed to identify the critical factors that drive the efficacy of our residual MLP model equipped with Jacobian regularization. By selectively disabling certain features, we aim to provide insights into the essential elements of our approach and their impact on learning outcomes. This helps in understanding the robustness and sensitivity of our model under different configurations.

In the first experiment, we conducted model ablations using three datasets: VAR100, fMRI, and Lorenz-96. We evaluated the impact of omitting the input-output Jacobian regularizer and varying the number of residual network layers on the performance of the algorithm. The results are detailed in Table \ref{tab:ablation}. The Jacobian matrix regularizer proved highly beneficial for capturing Granger causality. The advantage of adding residual layers was marginal for the linear VAR100 dataset but particularly advantageous for the nonlinear Lorenz-96 dataset. These findings confirm the versatility and efficacy of our approach across different scenarios and underscore the importance of tailoring the model to the specific characteristics of each dataset.

In the second experiment, we compare the impact of employing sparsity constraints versus Directed Acyclic Graph (DAG) constraints on the input-output Jacobian matrix using the VAR(100, 5, 10) and DREAM4 dataset \cite{marbach2009generating}.
Table \ref{tab:ablation-study} indicates that our approach is robust to the choice of ground-truth structure and performs better than the DAG regularizer on the input-output Jacobian matrix.

\subsection{Alternative neural architecture.}
Our Jacobian regularizer functions independently of the network structure. We chose the residual-MLP architecture for its simplicity and efficiency. To show the regularizer’s structure-agnostic nature, we conducted two experiments using the Lorenz-96 dataset with a forcing parameter of $F=40$. The first experiment assessed the detection of summary causal graphs with a cLSTM model, a standard LSTM, and an LSTM enhanced by our Jacobian regularizer. Notably, the cLSTM model only identified the summary causal graph. The second experiment evaluated full-time causal relationships using both the standard LSTM and the enhanced LSTM. As shown in Table \ref{tab:anotherar}, the results confirm that our Jacobian regularizer is independent of the network's structure and effectively improves the model’s ability to learn both summary and full-time Granger causality.
More details can be seen in the appendix.

\subsection{Causal discovery and time series forecasting}

As the definition of Granger causality, its assessment is based on the performance of prediction. 
A better prediction model and a better Granger causality constraint will help each other. 
In other words, we believe that the Granger causality constraint will help the model avoid overfitting the data.
We take an example of the Lorenz-96 dataset with $F=40$ and see if the predictive performance will be enhanced by introducing the Granger causality constraint, i.e., the input-output Jacobian regularizer.
Specifically, we compare the training and testing losses with and without Jacobian constraints. We also evaluate the mean squared error (MSE) between the predicted value and true value under both constrained and unconstrained conditions.
The results depicted in Table \ref{tab:lorenzf40mse1} highlight how Granger causality contributes to enhancing the generalization capability of the predictive model and the causal constraint (i.e., input-output Jacobian regularizer) helps reduce overfitting and improve prediction performance.

\section{Conclusion}
In this work, we introduce a novel neural Granger causality method, termed Jacobian Regularizer-based Neural Granger Causality (JRNGC), which incorporates a new framework and an input-output Jacobian regularizer.
By representing Granger causality through the input-output Jacobian matrix, our method eliminates the need to build separate models for each target variable and the sparsity on the weights of the first layer. Specifically, we utilize a joint model and apply a sparsity penalty to the input-output Jacobian matrix.
Extensive experiments demonstrate that our method enhances the modeling of interactions within time series data and enables a more accurate assessment of both full-time Granger causality and summary Granger causality.
Our code is available at https://github.com/ElleZWQ/JRNGC.

\section*{Impact Statement}
This paper presents work whose goal is to advance the field of 
Machine Learning. There are many potential societal consequences 
of our work, none of which we feel must be specifically highlighted here.

\section*{Acknowledgments}
\label{sec:ack}
This work was supported by the National Natural Science Foundation of China with grant numbers (U21A20485, 62088102).


\bibliography{ref}
\bibliographystyle{icml2024}

\newpage
\onecolumn
\appendix
\twocolumn

\section{Background}
\subsection{Post-hoc analysis}

Post-hoc analysis is a commonly used phase in deep learning after training a neural network. 
For instance, scientists often utilize heat maps to determine which features are more effective for classification.
Similarly, time series forecasting models are designed to predict future values but not to explicitly capture Granger causality. Therefore, a post-hoc phase is necessary for Granger causality analysis. 
In our approach, the post-hoc phase can be summarized as follows: after training the multivariate time series forecasting model with Jacobian regularizer, our objective is to explore Granger causality across various time lags. To achieve this, we conduct significant tests on the learned input-output Jacobian matrix.  The significance of the elements in the input-output Jacobian matrix determines the presence or absence of Granger causal relationships between variables. 
Specifically, if the value of an element in the Jacobian matrix is greater than zero, it indicates a Granger causal relationship between the variable corresponding to the row and the variable corresponding to the column. Conversely, if the value is zero or less than zero, it suggests no Granger causal relationship between the variables. The post-hoc analysis allows us to identify and quantify the Granger causality between different variables, providing valuable insights into the underlying dynamics of the multivariate time series. 
\subsection{Causal graphs for time series} 
\begin{figure}[htbp]
	\centering
	\includegraphics[width=0.47\textwidth]{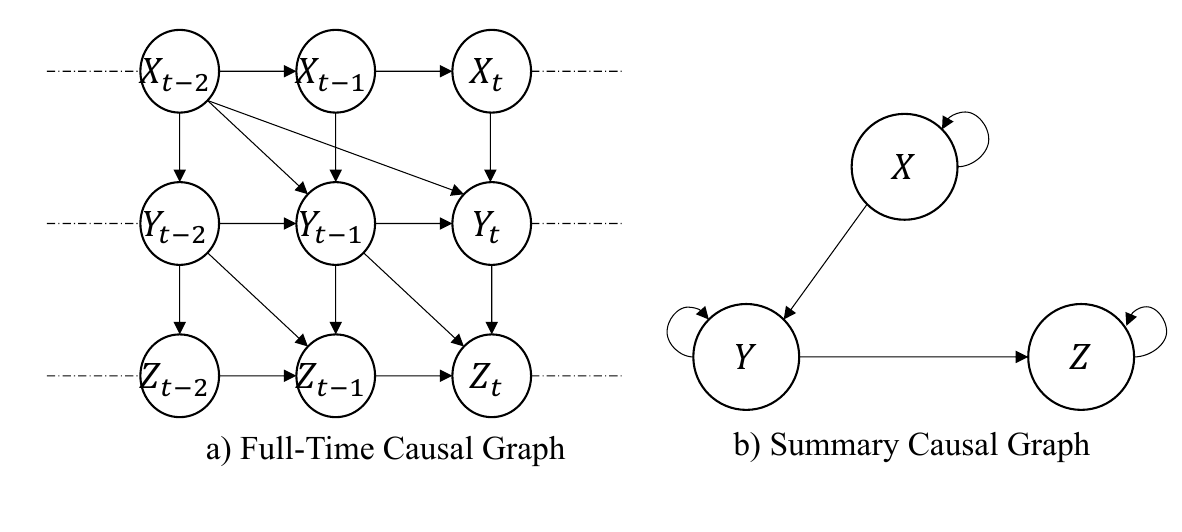}
	\caption{
		Typical causal graphs for time series. The full-time causal graph entails including all intricate causal links and interactions among variables. In contrast, the summary causal graph streamlines this complexity, focusing on the most significant causal connections.
	}
	\label{fig:causalgraph}
\end{figure}

Full-time causal graph and summary causal graph are always used for representing the causal relationships for time series (see Figure \ref{fig:causalgraph}).
It is always difficult to find the full-time causal graph, especially for the methods that use Long Short-Term Memory (LSTM) network \cite{tank2021neural}

\subsection{AUPRC: an important metric for learning sparse causality}

Sparse causality refers to situations where only a small subset of variables in a dataset actually exhibit causal relationships, while the majority remain unrelated. 
Variable usage percentage is a measure that quantifies the proportion of variables involved in causal relationships compared to the total number of variables.
In our context, a low variable usage percentage signifies a sparse causal structure. For instance, if only 14.67\% of variables are actively participating in causal relationships, it indicates a high degree of sparsity in the dataset.
In the context of learning sparse causality, AUPRC is more informative than AUROC. A model with a high AUROC score could simply be good at identifying unrelated variable pairs while performing poorly in identifying the few relevant causal relationships among the sparse variables. On the other hand, a high AUPRC score indicates that the model is proficient in correctly identifying and prioritizing the sparse causal relationships. Therefore, AUPRC is a better indicator of the model's ability to capture and learn the sparse causal structure, making it a crucial metric for evaluating the performance of methods focused on sparse causality.

\section{Datasets Details}

\textbf{VAR model.} Vector autoregressions (VAR) model is defined as:
\begin{equation}
	\mathbf{x}(t)=\sum_{\alpha=1}^\tau A_\alpha \mathbf{x}(t-\alpha)+\bm{\varepsilon},
\end{equation}
where $\mathbf{x}$ is $D$-dimensional time series with time index $t$ and $\tau$ is the maximum true time lag.  
$A_\alpha$ is the relation matrix between the $D$ time series at time $\alpha$. $\bm{\varepsilon}$ is a noise item.
In our experiments on VAR datasets, we denote VAR(100, 5, 10) to represent a scenario where there are 100 dimensions in total. Among these, the true time lag $\tau$ is set to 5, while the maximum estimated lag $\eta$ is set to 10.

\textbf{Lorenz-96.} It is a classic chaotic dynamical model, which describes the non-linear interaction between variables in systems such as atmospheric circulation and other natural phenomena \cite{karimi2010extensive}, which is defined as:
\begin{equation}
	\frac{d x_i(t)}{d t}=\left(x_{i+1}(t)-x_{i-2}(t)\right) x_{i-1}(t)-x_i(t)+F,
	\label{Lorenz-96}
\end{equation}
where $t$ is the time index and the constant $F$ is an important parameter representing the external forcing on the system.
The sequence index $i$ is taken modulo, i.e., $x_{-1} = x_{D-1}, x_{0}=x_{D} $ and $i = 1,2,3,\cdots, D$.

\textbf{fMRI data.} It is usually used to estimate the brain network. 
Stephen M. Smith et al. generated rich, realistic simulated fMRI data for a wide range of underlying networks, experimental scenarios, and problematic confounders in the data to compare different approaches to connectivity estimation \cite{smith2011network}. 
The dataset consists of 50 subjects, with each subject having 15 nodes and 200 observations.

\textbf{DREAM-3 dataset.} It is a realistic gene expression data set from the DREAM-3 challenge \cite{prill2010towards}.
This challenge includes five simulated datasets, comprising two E. coli datasets and three yeast datasets, each featuring a distinct underlying Granger causality plot. 
Each dataset contains $100$ numbers of different time series, each with 46 replicates, sampled at 21 time points, resulting in a total of 966 time points. 
As can be seen, the data is very limited in length and is a difficult non-linear dataset.

\textbf{DREAM-4 dataset.} The DREAM-4 network inference challenge, established by \cite{marbach2009generating}, aims to facilitate the learning of gene regulatory networks from gene expression data. The challenge comprises 5 independent datasets, each containing data from 10 different time-series recordings. These recordings capture the expression levels of 100 genes over 21 time steps. 

\textbf{CausalTime.} CausalTime is proposed by \cite{cheng2024causaltime} to evaluate time-series causal discovery algorithms in real applications.
\section{Model complexity}
While our method may not require the same level of attention as large-scale models in terms of parameter count, our contribution lies in successfully addressing the challenge of joint model learning in exploring Granger causal relationships among multiple time series without increasing the complexity of a single model.
We could exploit the same model or a simpler but more effective model.
As evidenced by our experimental results, our approach allows us to utilize a single model to investigate Granger causal relationships among multiple time series. 
Specifically, we analyze scenarios involving both 100-dimensional variables and 10-dimensional variables to compare our methods with other approaches.
The superiority of our algorithm can be observed from various perspectives, including experimental time, model parameter count, AUPRC, and AUROC, as shown in Table \ref{tab:time_param_var10} and \ref{tab:time_param_var_100}.
\section{Motivation}
To articulate our motivation clearly, we employ a Q-A format to address a key inquiry.

Q: Why should we use the single neural network (NN) for Granger causality analysis?

A: The justification for employing a single NN model with shared hidden layers in the neural Granger causality framework is multifaceted. Firstly, this approach represents a pioneering effort in the field. To the best of our knowledge, it is the inaugural application of such a model configuration for multivariate Granger causality analysis. The innovative design of this model allows for an integrated processing of time series data, which is critical given the complex interdependencies among variables as illustrated in Figure \ref{fig:causalgraph}.
Conventional frameworks that apply Granger causality assumptions or regularizers directly to the weights of only the first or last layer tend to overemphasize the importance of these weights. This not only distorts the model’s ability to accurately discern true causal relationships but also inadvertently magnifies the influence of non-causal variables in predictions. This misalignment can lead to incorrect conclusions about Granger causality, considering Granger causality's definition relies on enhancing prediction accuracy by including potentially causal variables' past values.
Additionally, the traditional approach of employing multiple models to predict multiple variables is inefficient. It necessitates training separate models for each variable, which is both resource-intensive and time-consuming. By contrast, our single NN model streamlines this process, enhancing operational efficiency without compromising the quality of causality analysis.
The performance metrics of our model, as demonstrated in our experimental results (see Table \ref{tab:time_param_var10} and Table \ref{tab:time_param_var_100}), attest to the effectiveness of using a unified framework. This approach not only simplifies the learning of causal relationships but also ensures a higher degree of accuracy in identifying and interpreting these relationships.

\section{Implementation Experiment Details}

\subsection{Significance test}

Considering the real scenes, we have no access to obtain the true causal relationships between variables. 
Instead, we aim to extract causal relationships from the estimated weighted causal matrix $\mathbf{W}$.
However, the learned causal relationships will inherently include noise, i.e., when there is no causal relationship between variables $i$ and $j$, $W_{i,j}$ or $W_{j,i}$ is challenging to be precisely equal to 0.
Therefore, we need to test whether each element of the weighted causal matrix is significantly larger than 0 \cite{zhou2022causality,suryadi2023granger}.

\begin{figure}[htbp]
\centering
\includegraphics[width=0.4\textwidth]{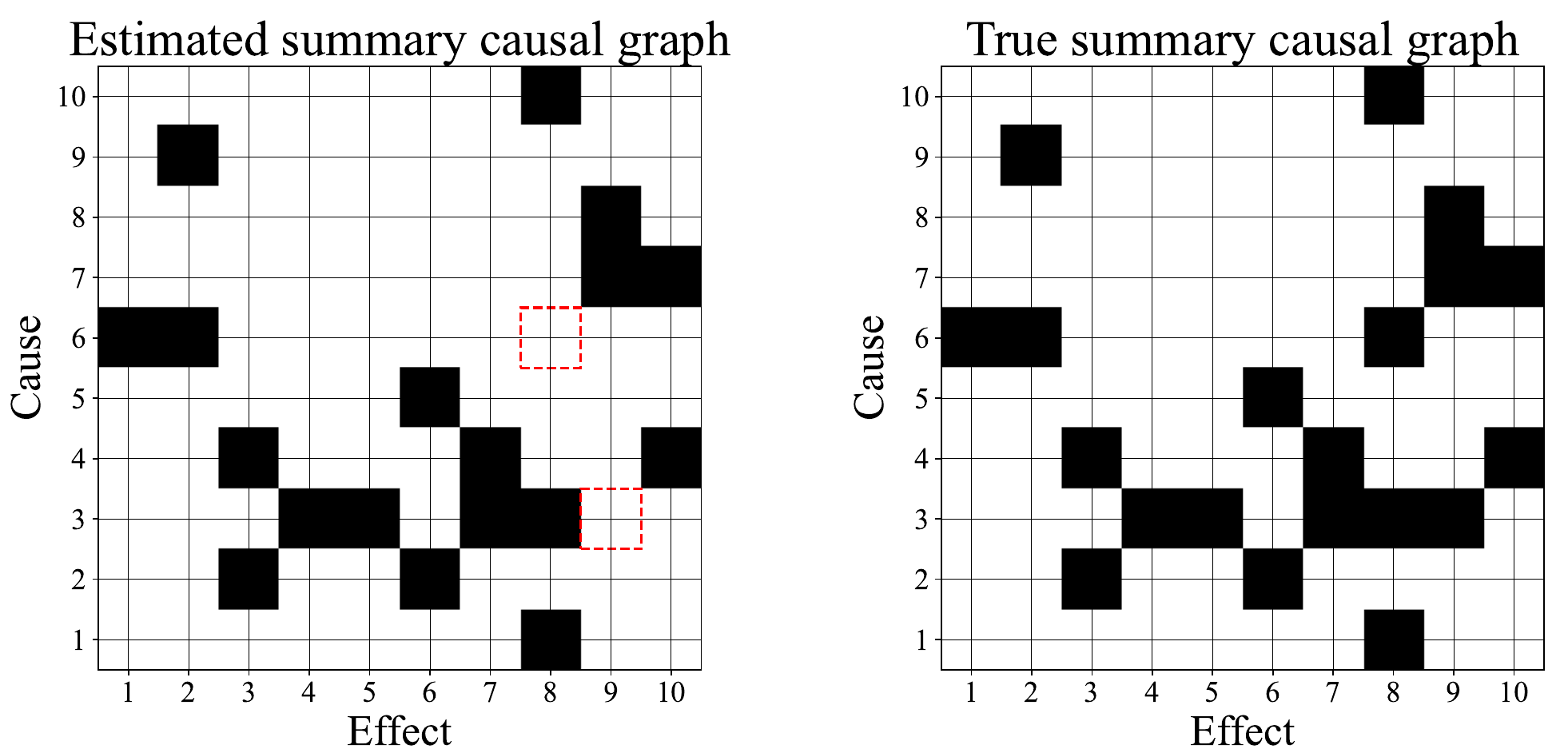}
\caption{Estimated summary Causal graph Results on VAR (10,3,5) dataset. 95\% confidence interval shown. }
\label{fig:summary-sig}
\end{figure}

\begin{figure}[htbp]
\centering
\includegraphics[width=0.4\textwidth]{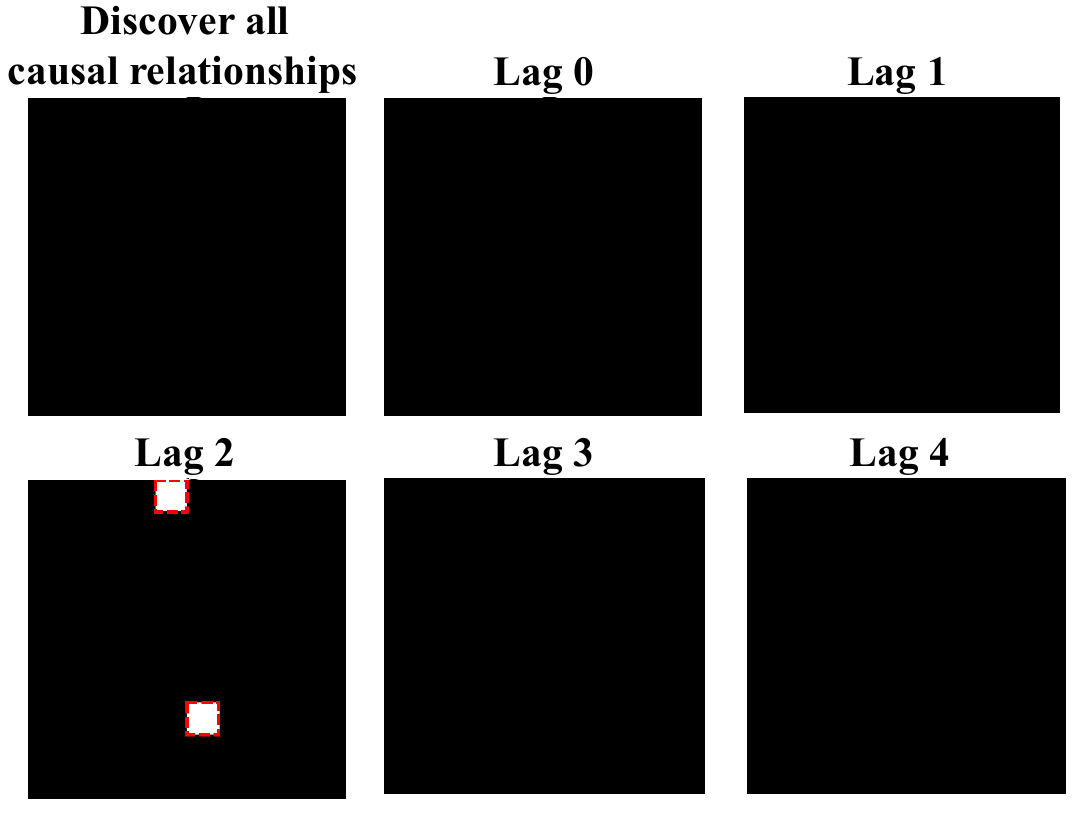}
\caption{Performance of full-time causal graph learning. Black denotes correctly identified causal relationships, and white denotes overlooked or disregarded relationships. Lag indicates the time $t-\text{lag}$. Our method can detect full-time causal relationships. 95\% confidence interval shown.}
\label{fig:full-time-sig}
\end{figure}

Here, we use the significance test method as  \cite{zhou2022causality}, i.e., generating surrogate datasets for hypothesis testing.
To generate surrogate datasets, we need to disrupt the intrinsic causal relationships within the data by eliminating the full-time interdependencies among variables.
For instance, let's consider three variables: $x_1$, $x_2$, and $x_3$, each containing 500 observations. To create surrogate datasets without causal links between them, we form the initial dataset by concatenating $x_1[0:100]$, $x_2[m:100+m]$, and $x_3[2m:100+2m]$, with $m$ set to a significantly large value.
Specifically, we set $m$ to 50, which surpasses the true lag (where the true lag value is 3), and the resulting dataset has a time length of 100. With our 10-dimensional VAR dataset having 600 observations, we generate 50 sets of surrogate datasets.
Training these surrogate datasets using the same model structure as the model for the original dataset, we derive the weighted causal matrix $\mathbf{Ws}$ for each surrogate dataset. As causal relationships are absent among variables in the surrogate datasets, we establish the threshold using the 3-sigma rule. Our threshold is defined as follows:

\begin{equation}
\begin{aligned}
\text{Ws}_{\mu} &= \frac{1}{\text{Num}_s}\sum\textbf{Ws}_i\\
\text{Ws}_{\sigma} &= \sqrt{\frac{1}{\text{Num}_s}\sum_{i=1}^{\text{Num}_s}(\textbf{Ws}_i - \text{Ws}_{\mu})^2},  \\
 \text{Thres} &=  \text{Ws}_{\mu} + 2 \text{Ws}_{\sigma}
\end{aligned}
\end{equation}
where $\textbf{Ws}_i$ is the weighted causal matrix for the $i$-th surrogate dataset.
$\text{Ws}_{\mu}$ is the mean of the weighted causal matrices.
$\text{Ws}_{\sigma}$ is the standard deviation of the weighted causal matrices, $\text{Num}_s$ is the number of surrogate datasets. 
As shown in Figure \ref{fig:summary-sig} and \ref{fig:full-time-sig}, we have good performance in obtaining the summary and full-time causal graphs. Our method can estimate the causal graph of each time (e.g., $t-lag$) successfully.

\subsection{More details on the experiment for DREAM-3 dataset}

We supplement the experimental results of NAVAR(MLP), NAVAR(LSTM), JGC, CR-VAE, and our methods on the DREAM-3 dataset. Table \ref{tab:dreamappen} shows that our approach demonstrates superior performance compared to other algorithms. 

\subsection{More details for experiment for alternative neural architecture}
In this experiment, the reason why the LSTM model can achieve full-time Granger causality is that we employed the same post-processing procedure as the one described in this paper. This approach was chosen to better highlight the effectiveness of our method.

\begin{table}[ht]
\small
\caption{Comparisons of AUROC and AUPRC on DREAM-3 dataset.}
\vskip 0.15in
\centering
\begin{tabular}{cccc}
\toprule
Dataset & Model & AUROC & AURPC \\
\midrule
\multirow{5}{*}{E.colo-1} & NAVAR(MLP) & 0.557 & 0.102 \\
     & NAVAR(LSTM) & 0.544 & 0.013 \\
     & JGC & 0.504 & 0.018 \\
     & CR-VAE & 0.502 & 0.013 \\
     & JRNGC-F & 0.666 & 0.198 \\
\midrule
\multirow{5}{*}{E.coli-2} & NAVAR(MLP) & 0.577 & 0.107 \\
     & NAVAR(LSTM) & 0.473 & 0.012 \\
     & JGC & 0.527 & 0.016 \\
     & CR-VAE & 0.494 & 0.012 \\
     & JRNGC-F & 0.678 & 0.202 \\
\midrule
\multirow{5}{*}{Yeast-1} & NAVAR(MLP) & 0.652 & 0.073 \\
     & NAVAR(LSTM) & 0.497 & 0.030 \\
     & JGC & 0.604 & 0.026 \\
     & CR-VAE & 0.525 & 0.017 \\
     & JRNGC-F & 0.650 & 0.172 \\
\midrule
\multirow{5}{*}{Yeast-2} & NAVAR(MLP) & 0.573 & 0.105 \\
     & NAVAR(LSTM) & 0.477 & 0.038 \\
     & JGC & 0.553 & 0.050 \\
     & CR-VAE & 0.518 & 0.040 \\
     & JRNGC-F & 0.597 & 0.142 \\
\midrule
\multirow{5}{*}{Yeast-3} & NAVAR(MLP) & 0.548 & 0.089 \\
     & NAVAR(LSTM) & 0.466 & 0.052 \\
     & JGC & 0.521 & 0.059 \\
     & CR-VAE & 0.501 & 0.055 \\
     & JRNGC-F & 0.560 & 0.130 \\
    \bottomrule
\end{tabular}
\label{tab:dreamappen}
\vskip 0.1in
\end{table}

\subsection{Performance on a larger dataset} 

Although dream3 and fMRI are benchmark datasets, they have limitations for NN methods.
We experimented with an epilepsy dataset, using 12 EEG time series selected from deeper brain structures with 4000 simulations. The dataset lacks causal ground truth. As shown in Fig \ref{fig:se}, a), the result aligns with clinical practice, where doctors often operate in the last six areas.

\subsection{Jacobian regularizer hyperparameters sensitivity} 
\begin{figure}[htbp]
  \centering
  \includegraphics[width=0.5\textwidth]{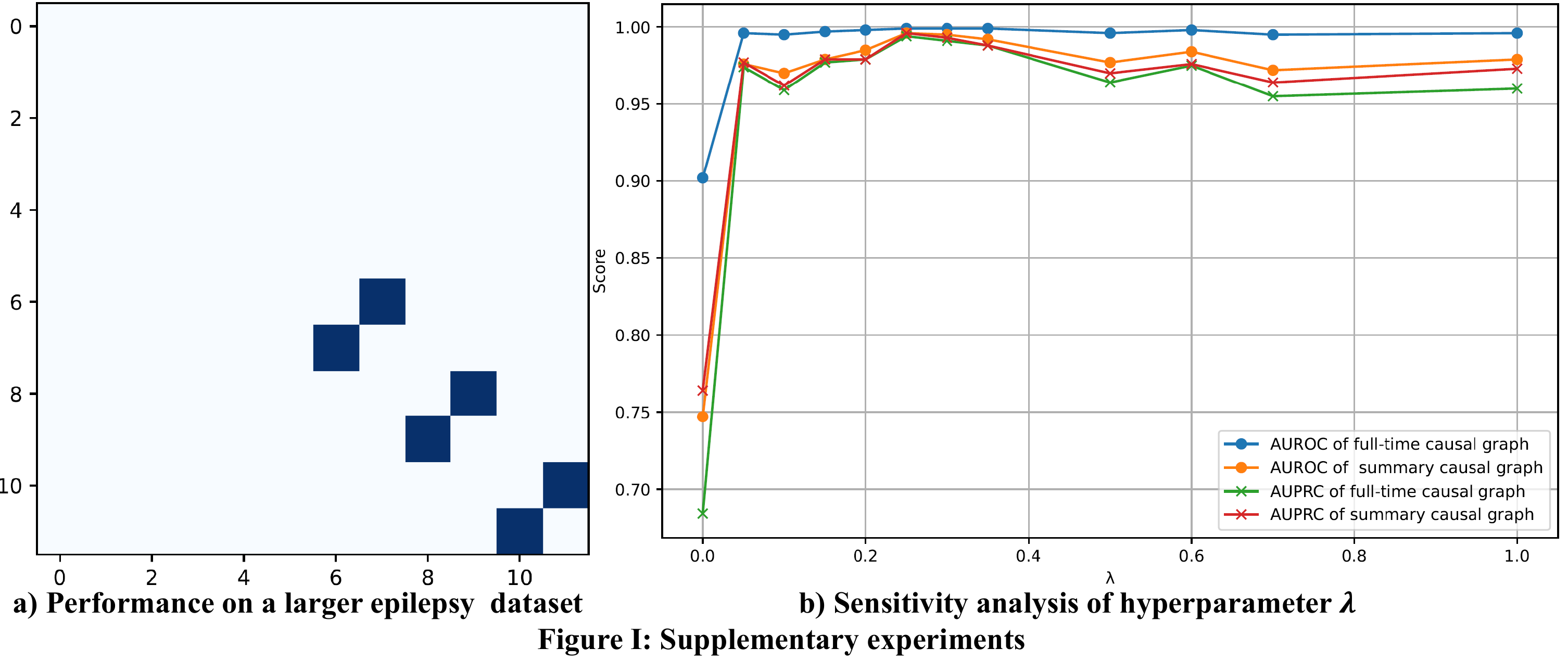}
  \caption{Granger Causality analysis by JRNGC on EEG dataset (left) and sensitivity analysis of $\lambda$ (right) on Lorenz dataset.}
  \label{fig:se}
\end{figure}

Figure \ref{fig:se} b) illustrates the outcomes of varying the parameter $\lambda$. The findings indicate that our method exhibits robust performance across a range of values for both parameters. Additionally, with regards to the sensitivity of the number of random projections, it has been observed in \cite{hoffman2019robust} that once the simulation length surpasses 100, the number of random projections has minimal impact on the results.

\onecolumn

\onecolumn

\begin{table*}[ht]
	\caption{Comparison of tunable parameters and training time per epoch on the Var10 dataset.}
 \vskip 0.15in
	\centering
	\begin{tabular}{lcccc}
		\toprule
		Model & Number of tunable Parameters & Training time of an epoch(s)  & AUROC ($\uparrow$) & AUPRC ($\uparrow$)  \\ 
		\midrule
		cMLP & 52010 & 0.028 &0.978\smallgrey{$\pm$0.032}& 0.923\smallgrey{$\pm$0.110}  \\ 
		cLSTM & 124510 & 0.031 & 0.931\smallgrey{$\pm$0.061} & 0.803\smallgrey{$\pm$0.162}  \\ 
		NVAR(MLP) & \underline{8110} & \underline{0.017} & 0.976\smallgrey{$\pm$0.131} & 0.936\smallgrey{$\pm$0.086} \\ 
		NVAR(LSTM) & 111110 & 0.131 & 0.952\smallgrey{$\pm$0.064} & 0.875\smallgrey{$\pm$0.156} \\ 
		JGC &56600 & 0.203  & \textbf{1.000}\smallgrey{$\pm$0.001}& \underline{0.998}\smallgrey{$\pm$0.003}  \\ 
		CR-VAE & 390810 & 0.027 &0.749\smallgrey{$\pm$0.013}&0.401\smallgrey{$\pm$0.107} \\ 
		\textbf{JRNGC-L1(ours)} & \textbf{3110} & 0.032 & \textbf{1.000}\smallgrey{$\pm$0.001}& \textbf{0.999}\smallgrey{$\pm$0.003}  \\ 
		\textbf{JRNGC-F(ours)} & \textbf{3110} & \textbf{0.005}& \underline{0.996}\smallgrey{$\pm$0.007}  &0.994\smallgrey{$\pm$0.011}  \\ 
		
		\bottomrule
	\end{tabular}
	\label{tab:time_param_var10}
 \vskip -0.1in
\end{table*}

\begin{table*}[htbp]
	 \caption{Comparison of tunable parameters and training time per epoch on the Var100 dataset.}
  \vskip 0.15in
	\centering
	\begin{tabular}{lcccc}
		\toprule
		Model & Number of tunable Parameters & Training time of an epoch(s) & AUROC ($\uparrow$)  & AUPRC ($\uparrow$) \\
		\midrule
		cMLP & 11030100 & 0.217 & 0.940\smallgrey{$\pm$0.0013} & 0.851\smallgrey{$\pm$0.051}  \\ 
		cLSTM & 15135100 & 0.433 &0.845\smallgrey{$\pm$0.045} & 0.606\smallgrey{$\pm$0.102}  \\ 
		NAVAR(MLP) & \underline{1120100} &\underline{0.175} & 0.887\smallgrey{$\pm$0.023} & 0.776\smallgrey{$\pm$0.040}  \\ 
		NAVAR(LSTM) & 5130100 & 0.691 & 0.860\smallgrey{$\pm$0.011}  & 0.750\smallgrey{$\pm$0.018}  \\ 
		JGC & 5870000 & 2.333 & \textbf{1.000}\smallgrey{$\pm$0.000} & \textbf{1.000}\smallgrey{$\pm$0.000} \\ 
		CR-VAE & 6150900 & 0.190 & 0.645\smallgrey{$\pm$0.014} &0.268\smallgrey{$\pm$0.008}  \\ 
		\textbf{JRNGC-L1(ours)} & \textbf{110300} & 9.799& \textbf{1.000}\smallgrey{$\pm$0.000} & \textbf{1.000}\smallgrey{$\pm$0.000}  \\ 
		\textbf{JRNGC-F(ours)} & \textbf{110300} & \textbf{0.033} & \underline{0.984}\smallgrey{$\pm$0.002} & \underline{0.960}\smallgrey{$\pm$0.005}  \\ 
		
		\bottomrule
	\end{tabular}
	\label{tab:time_param_var_100}
  \vskip -0.1in
\end{table*}

\section{Experimental Hyperparameters}
Here, we present the tuned hyperparameters for our methods and the comparative approaches across various datasets in our experiments from Table \ref{cMLP_var} to Table \ref{causaltime}.

\begin{table*}[htbp]
\caption{Tuned Hyperparameters of cMLP on the VAR Dataset.}
\vskip 0.15in
\centering
\resizebox{\textwidth}{!}{
\begin{tabular}{lcccccc}
\toprule
~ & Hidden size & gate\_lam   & gate\_regular &regular\_lam  & regular\_type & Learning rate  \\ \midrule
Tuning range   & [50,100] & [0.1,1] & ['EL', 'GL', 'GSGL', 'H', 'SPH']  &[0.01] & ['lasso', 'ridge'] & [5e-3,1e-1]    \\ \midrule
VAR ($10,3,5$) & [100] & 0.1  & 'H' & 0.01 & 'ridge' & 5e-2 \\
VAR ($50,5,10$) & [100] & 0.01  & 'H' & 0.01 & 'ridge' & 5e-2 \\    
VAR ($100,5,10$) & [100,100] & 0.001  & 'GL' & 0.01 & 'ridge' & 5e-2 \\    
\bottomrule   
\end{tabular}
}

\label{cMLP_var}
\vskip -0.1in
\end{table*}

\begin{table*}[htbp]
\caption{Tuned Hyperparameters of cMLP on the Lorenz-96 ($F= 10, 40$) Dataset.}
\vskip 0.15in
\centering
\resizebox{\textwidth}{!}{
\begin{tabular}{lcccccc}
\toprule
~ & Hidden size & gate\_lam   & gate\_regular &regular\_lam  & regular\_type & Learning rate  \\ \midrule
Tuning range   & [50,100] & [0.1,1] & ['EL', 'GL', 'GSGL', 'H', 'SPH']  &[0.01] & ['lasso', 'ridge'] & [5e-3,1e-1]    \\ \midrule
Lorenz-96 ($F=10$) & [100] & 0.1  & 'H' & 0.01 & 'ridge' & 5e-2 \\
Lorenz-96 ($F=40$) & [100] & 0.1  & 'H' & 0.01 & 'ridge' & 1e-2 \\         
\bottomrule   
\end{tabular}
}
\label{cMLP_lorenz}
\vskip -0.1in
\end{table*}

\begin{table*}[htbp]
\caption{Tuned Hyperparameters of cMLP on the fMRI Dataset.}
\vskip 0.15in
\centering
\resizebox{\textwidth}{!}{
\begin{tabular}{lcccccc}
\toprule
~ & Hidden size & gate\_lam   & gate\_regular &regular\_lam  & regular\_type & Learning rate  \\ \midrule
Tuning range   & [50,100] & [0.1,1] & ['EL', 'GL', 'GSGL', 'H', 'SPH']  &[0.01] & ['lasso', 'ridge'] & [5e-3,1e-1]    \\ \midrule
~ & [50] & 0.1  & 'GL' & 0.01 & 'ridge' & 5e-2 \\
\bottomrule
\end{tabular}
}
\label{cMLP_fmri}
\vskip -0.1in
\end{table*}

\begin{table*}[htbp]
\caption{Tuned Hyperparameters of cLSTM on VAR Dataset.}
\vskip 0.15in
\centering
\begin{tabular}{lcccccc}
\toprule
~ & Hidden size & gate\_lam   & gate\_regular &regular\_lam   & Learning rate  \\ \midrule
Tuning range   & [50,100] & [0.1,1] & ['EL', 'GL']  &[0.01] & [5e-3,1e-1]    \\ \midrule
VAR ($10,3,5$) & 50 & 0.1  & 'GL' & 0.01  & 5e-2 \\
VAR ($50,5,10$) & 100 & 0.01  & 'GL' & 0.01 & 5e-3 \\ 
VAR ($50,5,10$) & 100 & 0.01  & 'GL' & 0.01 & 5e-3 \\ 
VAR ($100,5,10$) & 150 & 0.01  & 'GL' & 0.01 & 5e-3 \\ 
\bottomrule
\end{tabular}
\label{cLSTM_var}
\vskip -0.1in
\end{table*}

\begin{table*}[H]
\caption{Tuned Hyperparameters of cLSTM on the Lorenz-96 ($F= 10, 40$) Datasets.}
\vskip 0.15in
\centering
\begin{tabular}{lcccccc}
\toprule
~ & Hidden size & gate\_lam   & gate\_regular &regular\_lam   & Learning rate  \\ \midrule
Tuning range   & [50,100] & [0.1,1] & ['EL', 'GL']  &[0.01] & [5e-3,1e-1]    \\ \midrule
Lorenz-96 ($F=10$) & 50 & 0.01  & 'GL' & 0.01  & 5e-2 \\
Lorenz-96 ($F=40$) & 50 & 0.01  & 'GL' & 0.01 & 5e-2 \\ 
\bottomrule
\end{tabular}
\label{cLSTM_lorenz}
\vskip -0.1in
\end{table*}

\begin{table*}[htbp]
\caption{Tuned Hyperparameters of cLSTM on the fMRI Datasets.}
\vskip 0.15in
\centering
\begin{tabular}{lcccccc}
\toprule
~ & Hidden size & gate\_lam   & gate\_regular &regular\_lam   & Learning rate  \\ \midrule
Tuning range   & [50,100] & [0.1,1] & ['EL', 'GL']  &[0.01] & [5e-3,1e-1]    \\ \midrule
~ & 50 & 0.1  & 'GL' & 0.01  & 5e-2 \\
\bottomrule
\end{tabular}
\label{cLSTM_fmri}
\vskip -0.1in
\end{table*}

\begin{table*}[htbp]
\caption{Tuned Hyperparameters of NAVAR (MLP) on the VAR Datasets. K is the Number of Lags Considered, $\lambda$ is the Contribution Penalty, and $\mu$ is the Weight Decay.}
\vskip 0.15in
\centering
\begin{tabular}{lccccccc}
\toprule
~ & K & Hidden Units & Layers   & Batch Size & Learning rate & $\lambda$ & $\mu$  \\ 
\midrule
Tuning range   & [5,10] & [50,100] &- &[32,128]& [5e-5,5e-3] & [0,1]  & [1e-7,0.5]   \\ 
\midrule
VAR ($10,3,5$) & 5 & 50 & 1 & 64 & 5e-4 & 0.02 & 9e-3   \\ 
VAR ($50,5,10$) & 10 & 100 & 1 & 64 & 5e-4 & 0.02 & 1e-4   \\ 
VAR ($100,5,10$) & 10 & 100 & 1 & 64 & 5e-4 & 0.02 & 1e-4   \\ 
\bottomrule
\end{tabular}
\label{navar_mlp}
\vskip -0.1in
\end{table*}

\begin{table*}[htbp]
\caption{Tuned Hyperparameters of NAVAR (MLP) on the Lorenz-96 ($F=10, 40$) Datasets. K is the Number of Lags Considered, $\lambda$ is the Contribution Penalty, and $\mu$ is the Weight Decay.}
\vskip 0.15in
	\centering
	\begin{tabular}{lccccccc}
		\toprule
		~ & K & Hidden Units & Layers   & Batch Size & Learning rate & $\lambda$ & $\mu$  \\ \midrule
		Tuning range   & - & - &- &-& [5e-5,5e-3] & [0,1]  & [1e-7,0.5]   \\ \midrule
		Lorenz-96 ($F=10$) & 5 & 50 & 1 & 64 & 5e-4 & 0.5 & 9e-3   \\ 
		Lorenz-96 ($F=40$) & 5 & 50 & 1 & 64 & 5e-3 & 0.4 & 9e-3   \\ 
		\bottomrule
	\end{tabular}
 \vskip -0.1in
\end{table*}

\begin{table*}[htbp]
\caption{Tuned Hyperparameters of NAVAR (MLP, LSTM) on the fMRI Datasets. K is the Number of Lags Considered, $\lambda$ is the Contribution Penalty, and $\mu$ is the Weight Decay.}
\vskip 0.15in
\centering
\begin{tabular}{lccccccc}
\toprule
Model & K & Hidden Units & Layers   & Batch Size & Learning rate & $\lambda$ & $\mu$  \\ \midrule
Tuning range   & - & - &- &-& [5e-5,5e-3] & [0,1]  & [1e-7,0.5]   \\ \midrule
NAVAR (MLP) & 5 & 50 & 1 & 64 & 1e-4 & 0.5 & 9e-3   \\ 
NAVAR (LSTM) & 5 & 50 & 1 & 64 & 1e-4 & 0.2 & 8e-4   \\ 
\bottomrule
\end{tabular}
\vskip -0.1in
\end{table*}

\begin{table*}[htbp]
\caption{Tuned Hyperparameters of NAVAR (MLP) on the DREAM-3 Datasets. K is the Number of Lags Considered, $\lambda$ is the Contribution Penalty, and $\mu$ is the Weight Decay. The tuned hyperparameters are copied from \cite{bussmann2021neural}.}
\vskip 0.15in
\centering
\begin{tabular}{lccccccc}
\toprule
~ & K & Hidden Units & Layers   & Batch Size & Learning rate & $\lambda$ & $\mu$  \\ \midrule
Tuning range   & - & - &[1,4] &[16,256]& [5e-5,5e-3] & [0,0.5]  & [1e-7,0.5]   \\ \midrule
Ecoli1 & 2 & 10 & 1 & 128 & 0.0005 & 0.1883 & 1.114e-4   \\ 
Ecoli2 & 2 & 10 & 1 & 32 & 0.001 & 0.2011 & 1.710e-4   \\ 
Yeast1 & 2 & 10 & 2 & 16 & 0.002 & 0.2697 & 1.424e-4   \\ 
Yeast2 & 2 & 10 & 1 & 256 & 0.0002 & 0.1563 & 2.013e-4   \\ 
Yeast3 & 2 & 10 & 1 & 16 & 0.0002 & 0.1559 & 1.644e-4 \\ \bottomrule
\end{tabular}
\vskip -0.1in
\end{table*}

\begin{table*}[htbp]
\caption{Tuned Hyperparameters of NAVAR (LSTM) on the VAR Datasets. K is the Number of Lags Considered, $\lambda$ is the Contribution Penalty, and $\mu$ is the Weight Decay.}
\vskip 0.15in
\centering
\begin{tabular}{lccccccc}
\toprule
~ & K & Hidden Units & Layers   & Batch Size & Learning rate & $\lambda$ & $\mu$  \\ \midrule
Tuning range   & [5,10] & [50,100] &- &[32,128]& [5e-5,5e-3] & [0,1]  & [1e-7,0.5]   \\ \midrule
VAR ($10,3,5$) & 5 & 50 & 1 & 128 & 1e-4 & 0.01 &7.5e-4   \\ 
VAR ($50,5,10$) & 10 & 100 & 1 & 128 & 1e-4 & 0.01 & 7e-5   \\ 
VAR ($100,5,10$) & 10 & 100 & 1 & 128 & 1e-4 & 0.04 & 7e-5   \\ 
\bottomrule
\end{tabular}
\vskip -0.1in
\end{table*}

\begin{table*}[htbp]
\caption{Tuned Hyperparameters of NAVAR (LSTM) on the Lorenz-96 ($F=10, 40$) Datasets. K is the Number of Lags Considered, $\lambda$ is the Contribution Penalty, and $\mu$ is the Weight Decay.}
\vskip 0.15in
\centering
\begin{tabular}{lccccccc}
\toprule
~ & K & Hidden Units & Layers   & Batch Size & Learning rate & $\lambda$ & $\mu$  \\ \midrule
Tuning range   & - & - &- &-& [5e-5,5e-3] & [0,1]  & [1e-7,0.5]   \\ \midrule
Lorenz-96 ($F=10$) & 5 & 50 & 1 & 64 &5e-4 & 0.01 & 3e-3   \\ 
Lorenz-96 ($F=40$) & 5 & 50 & 1 & 64 & 1e-4 & 0.078 & 8e-4   \\ 
\bottomrule
\end{tabular}
\vskip -0.1in
\end{table*}

\begin{table*}[htbp]
\caption{Tuned Hyperparameters of NAVAR (LSTM) on the DREAM-3 Datasets. K is the Number of Lags Considered, $\lambda$ is the Contribution Penalty, and $\mu$ is the Weight Decay. The tuned hyperparameters are copied from \cite{bussmann2021neural}.}
\vskip 0.15in
\centering
\begin{tabular}{lccccccc}
\toprule
~ & K & Hidden Units & Layers   & Batch Size & Learning rate & $\lambda$ & $\mu$  \\ \midrule
Tuning range   & - & - &- &-& [5e-5,5e-3] & [0,0.5]  & [1e-7,0.5]   \\ \midrule
Ecoli1 & 21 & 10 & 1 & 46 & 0.002 & 0.2208 & 1.094e-5   \\ 
Ecoli2 & 21 & 10 & 1 & 46 & 0.002 & 0.1958 & 3.233e-6   \\ 
Yeast1 & 21 & 10 & 1 & 46 & 0.002 & 0.2343 & 5.309e-5   \\ 
Yeast2 & 21 & 10 & 1 & 46 & 0.002 & 0.2189 & 1.987e-5   \\ 
Yeast3 & 21 & 10 & 1 & 46 & 0.002 & 0.2128 & 1.049e-5 \\ \bottomrule
\end{tabular}
\vskip -0.1in
\end{table*}

\begin{table*}[htbp]
\caption{Hyperparameters of JGC on the VAR, Lorenz-96, fMRI, DREAM-3 . The hyperparameters are copied from the supplementary materials of the method JGC \cite{suryadi2023granger} }
\vskip 0.15in
\centering
\resizebox{\textwidth}{!}{
\begin{tabular}{lcccccccc}
\toprule
DATASET & MAX LAG & HIDDEN LAYERS   & HIDDEN UNITS &TRAINING EPOCHS  & LEARNING RATE & BATCH SIZE & SPARSITY HYPERPARAMS  \\ \midrule
VAR & [5 10] & 2 & 50  &2000 & 1e-3& 64 & $\lambda \in [0.5,2.5]$   \\ 
Lorenz96  & 5 & 2  & 50 & 2000 &1e-3 &64 & $\lambda_{F=10} \in [0.5,2.5] \lambda_{F=40} \in [0.5,2.5]$ \\
fMRI & 1 & 2  & 50 & 2000 & 1e-3 & 64 & $\lambda \in [1,5]$ \\ 
DREAM-3 & 10 & 2 & 50 & 2000 &1e-3 & 64 & $\lambda \in [0.5,5]$ \\
\bottomrule
\end{tabular}
}
\label{JGC_datasets}
\vskip -0.1in
\end{table*}

\begin{table*}[htbp]
\caption{Tuned Hyperparameters of CR-VAE. K is the Number of Lags Considered, $\lambda$ is the sparsity-inducing penalty. $\lambda$-ridge is the ridge penalty at linear layer and hidden-hidden weights. Include\_self is used for post-hoc analysis, i.e., generating the Granger weighted causal matrix.}
\vskip 0.15in
\centering
\begin{tabular}{lcccccc}
\toprule
~ & K & Hidden size   & Learning rate & $\lambda$ & $\lambda$-ridge & Include\_self  \\ \midrule
Tuning range   & [2,10] & [16,256] &[0.001,0.1] &[0.01,10]& [0,10] & [True False]   \\ \midrule
VAR ($10,3,5$) & 5 & 100 & 0.1 & 0.1 & 0 & False  \\ 
VAR ($50,5,10$) & 10 & 100 & 0.1 &0.15& 0 & False  \\ 
VAR ($100,5,10$) & 10 & 100 & 0.05 & 0.1 & 0 &False  \\ 
\midrule
Lorenz-96 ($F=10$) & 5 & 100  & 0.05 & 0.1 &0 & True \\
Lorenz-96 ($F=40$) & 5 & 100  & 0.1 & 0.05 &0 & True \\ 
\midrule
fMRI  & 5 & 256  & 0.1 & 0.05 &0 & False\\
\midrule
DREAM-3 & 2 & 100  & 0.1 & 0.005 &0.1 & True \\
\bottomrule
\end{tabular}
\vskip -0.1in
\end{table*}

\begin{table*}[htbp]
\caption{Tuned Hyperparameters of JRNGC-L1 on the VAR Datasets.}
\vskip 0.15in
\label{jrngc_l1_var}
\centering
\begin{tabular}{lccccc}
\toprule
~ & Hidden size & Dropout rate & Residual Layers  & Jacobian lam & Learning rate  \\ \midrule
Tuning range   & [50,100] & - &[0,5] & [1e-5,1] & [1e-5,1e-3]    \\ \midrule
VAR (10,3,5) & 50 & - & 0 & 0.01 & 1e-3   \\ 
VAR (50,5,10) & 100 & - & 0 & 0.001 & 1e-3 \\
VAR(100,5,10) & 100 & - & 0 & 0.0001 & 1e-3\\
\bottomrule
\end{tabular}
\vskip -0.1in
\end{table*}

\begin{table*}[htbp]
\caption{Tuned Hyperparameters of JRNGC-L1 on the Lorenz-96 ($F= 10, 40$) Datasets.}
\vskip 0.15in
\centering
\begin{tabular}{lccccc}
\toprule
~ & Hidden size & Dropout rate & Residual Layers  & Jacobian lam & Learning rate  \\ \midrule
Tuning range   & [50,100] & - &[0,5] & [1e-5,1] & [1e-5,1e-3]    \\ \midrule
Lorenz-96 ($F=10$) & 100 & 0.2 & 5 & 0.00001 & 1e-3 \\
Lorenz-96 ($F=40$) & 100 & 0.2 & 5 & 0.02 & 1e-5   \\ 
\bottomrule
\end{tabular}
\label{jrngc_l1_lorenz}
\vskip -0.1in
\end{table*}

\begin{table*}[htbp]
\caption{Tuned Hyperparameters of JRNGC on the fMRI Datasets.}
\vskip 0.15in
\centering
\begin{tabular}{lccccc}
\toprule
Model & Hidden size & Dropout rate & Residual Layers  & Jacobian lam & Learning rate  \\ \midrule
Tuning range   & [50,100] & - &[0,5] & [1e-5,1] & [1e-5,1e-3]    \\ \midrule
JRNGC-L1 & 100 & - & 0 & 0.01 & 1e-3   \\ 
JRNGC-F & 100 & -& 0& 1 & 2e-5 \\
\bottomrule
\end{tabular}
\label{jrngc_f_fmri}
\vskip -0.1in
\end{table*}

\begin{table*}[htbp]
\caption{Tuned Hyperparameters of JRNGC-F on the VAR Datasets.}
\vskip 0.15in
\centering
\begin{tabular}{lccccc}
\toprule
~ & Hidden size & Dropout rate & Residual Layers  & Jacobian lam & Learning rate  \\ 
\midrule
Tuning range   & [50,100] & - &[0,5] & [1e-5,1] & [1e-5,1e-3]    \\ 
\midrule
VAR (10,3,5) & 50 & - & 0 & 0.01 & 1e-3   \\ 
VAR (50,5,10) & 50 & 0.2 & 5 & 0.0001 & 1e-3 \\
VAR(100,5,10) & 100 & - & 0 & 0.01 & 1e-3\\
\bottomrule
\end{tabular}
\label{jrngc_f_var}
\vskip -0.1in
\end{table*}

\begin{table*}[htbp]
\caption{Tuned Hyperparameters of JRNGC-F on the Lorenz-96 ($F= 10, 40$) Datasets.}
\vskip 0.15in
\centering
\begin{tabular}{lccccc}
\toprule
~ & Hidden size & Dropout rate & Residual Layers  & Jacobian lam & Learning rate  \\ \midrule
Tuning range   & [50,100] & - &[0,5] & [1e-5,1] & [1e-5,1e-3]    \\ \midrule
Lorenz-96 ($F=10$) & 100 & 0.2 & 5 & 0.0001 & 1e-3 \\
Lorenz-96 ($F=40$) & 100 & 0.2 & 5 & 0.2 & 1e-5   \\ 
\bottomrule
\end{tabular}
\label{jrngc_f_lorenz}
\vskip -0.1in
\end{table*}

\begin{table*}[htbp]
\caption{Tuned Hyperparameters of JRNGC on the DREAM-3 Datasets.}
\vskip 0.15in
\centering
\begin{tabular}{cccccc}
\toprule
Model & Hidden size & Dropout rate & Residual Layers  & Jacobian lam & Learning rate  \\ 
\midrule
Tuning range   & [50,100] & - &[0,5] & [1e-5,1] & [1e-5,1e-3]    \\ 
\midrule
JRNGC-F & 100 & 0.2& 1& 0.001 &1e-3 \\
\bottomrule
\end{tabular}
\label{jrngc_f_dream3}
\vskip -0.1in
\end{table*}

\begin{table*}[htbp]
\caption{Tuned Hyperparameters of JRNGC on the DREAM-4 Datasets.}
\vskip 0.15in
\centering
\begin{tabular}{cccccc}
\toprule
Model & Hidden size & Dropout rate & Residual Layers  & Jacobian lam & Learning rate  \\ 
\midrule
Tuning range   & [50,100] & - &[0,5] & [1e-5,1] & [1e-5,1e-3]    \\ 
\midrule
JRNGC-F & 100 & 0.2& 0& 1 &1e-3 \\
\bottomrule
\end{tabular}
\label{jrngc_f_dream4}
\vskip -0.1in
\end{table*}

\begin{table*}[htbp]
\caption{Tuned Hyperparameters of JRNGC-F on the CausalTime Datasets.}
\vskip 0.15in
\centering
\begin{tabular}{lccccc}
\toprule
~ & Hidden size & Dropout rate & Residual Layers  & Jacobian lam & Learning rate  \\ \midrule
Tuning range   & [50,100] & - &[0,5] & [1e-5,1] & [1e-5,1e-3]    \\ \midrule
AQI & 50 & 0.2 & 0 & 0.002 & 2e-3 \\
Traffic  & 50 & 0.2 & 5 & 0.001 & 1e-3   \\ 
Medical & 100 & 0.2 & 0 & 0.022 & 1e-3 \\
\bottomrule
\end{tabular}
\label{causaltime}
\vskip -0.1in
\end{table*}

\newpage

\end{document}